\documentclass[lettersize,journal]{IEEEtran}
\usepackage{amsmath,amsfonts}
\usepackage{algorithmic}
\usepackage[linesnumbered,ruled,vlined]{algorithm2e}
\usepackage{array}
\usepackage[caption=false,font=normalsize,labelfont=sf,textfont=sf]{subfig}
\usepackage{textcomp}
\usepackage{stfloats}
\usepackage{url}
\usepackage{verbatim}
\usepackage{graphicx}
\usepackage{makecell}
\usepackage{booktabs} %% tables with three lines
\usepackage{multirow}
\usepackage{threeparttable}
\usepackage{cite}
\begin{document}

\title{Enhancing Regrasping Efficiency Using Prior Grasping Perceptions with Soft Fingertips}

\author{Qiyin Huang, Ruomin Sui, Lunwei Zhang, Yenhang Zhou, Tiemin Li and Yao Jiang,~\IEEEmembership{Member,~IEEE}
        % <-this % stops a space
\thanks{This work was supported in part by the National Natural Science Foundation of China under Grant 52375017; in part by the National Natural Science Foundation of China under Grant 52175017; and in part by the Joint Fund of Advanced Aerospace Manufacturing Technology Research under Grant U2017202.}% <-this % stops a space
\thanks{Qiyin Huang, Ruomin Sui, Lunwei Zhang, Tiemin Li, and Yao Jiang are with the Institute of Manufacturing Engineering, Department of Mechanical Engineering, Tsinghua University, Beijing 100084, China
        ({e-mail: huangqy21@mails.tsinghua.edu.cn; ruomins@foxmail.com; zlw21@mails.tsinghua.edu.cn; zhouyanh23@mails.tsinghua.edu.cn; litm@mail.tsinghua.edu.cn; jiangyaonju@126.com}).}%
}

%\markboth{IEEE TRANSACTIONS ON ROBOTICS}%

\IEEEpubid{0000--0000\copyright~2024 IEEE.}
\IEEEpubidadjcol

\maketitle

\begin{abstract} % 需要后续抓取不能从否定第一次抓取出发。需要单纯从需要后续抓取出发。实验需要多批次物体（五类水果，同一类水果多个物体）。修改图片，第二次抓取改变位置或者改变物体，不能说同一物体的重复抓取。
Grasping the same object in different postures is often necessary, especially when handling tools or stacked items. Due to unknown object properties and changes in grasping posture, the required grasping force is uncertain and variable. Traditional methods rely on real-time feedback to control the grasping force cautiously, aiming to prevent slipping or damage. However, they overlook reusable information from the initial grasp, treating subsequent regrasping attempts as if they were the first, which significantly reduces efficiency. To improve this, we propose a method that utilizes perception from prior grasping attempts to predict the required grasping force, even with changes in position. We also introduce a calculation method that accounts for fingertip softness and object asymmetry. Theoretical analyses demonstrate the feasibility of predicting grasping forces across various postures after a single grasp. Experimental verifications attest to the accuracy and adaptability of our prediction method. Furthermore, results show that incorporating the predicted grasping force into feedback-based approaches significantly enhances grasping efficiency across a range of everyday objects.
\end{abstract}


\section{Introduction}
% 全文时态、语态
\IEEEPARstart{T}{he} ability to grasp objects is crucial for enabling a wide range of in-hand manipulations. A successful grasp requires applying the right amount of force to lift an object off the ground without causing slippage between the fingers and the object \cite{rolands.2009Coding}. In situations where the applied force is insufficient, the fingers may fail to provide the required force, leading to potential object slippage and unintended drops. On the other hand, excessive force can result in unwanted deformation or even damage, particularly with delicate items such as fruits and vegetables \cite{baohua2020Stateoftheart}, \cite{rinto2023Learning}.

\begin{figure}[t]
    \centering
    \includegraphics[width = 1\linewidth]{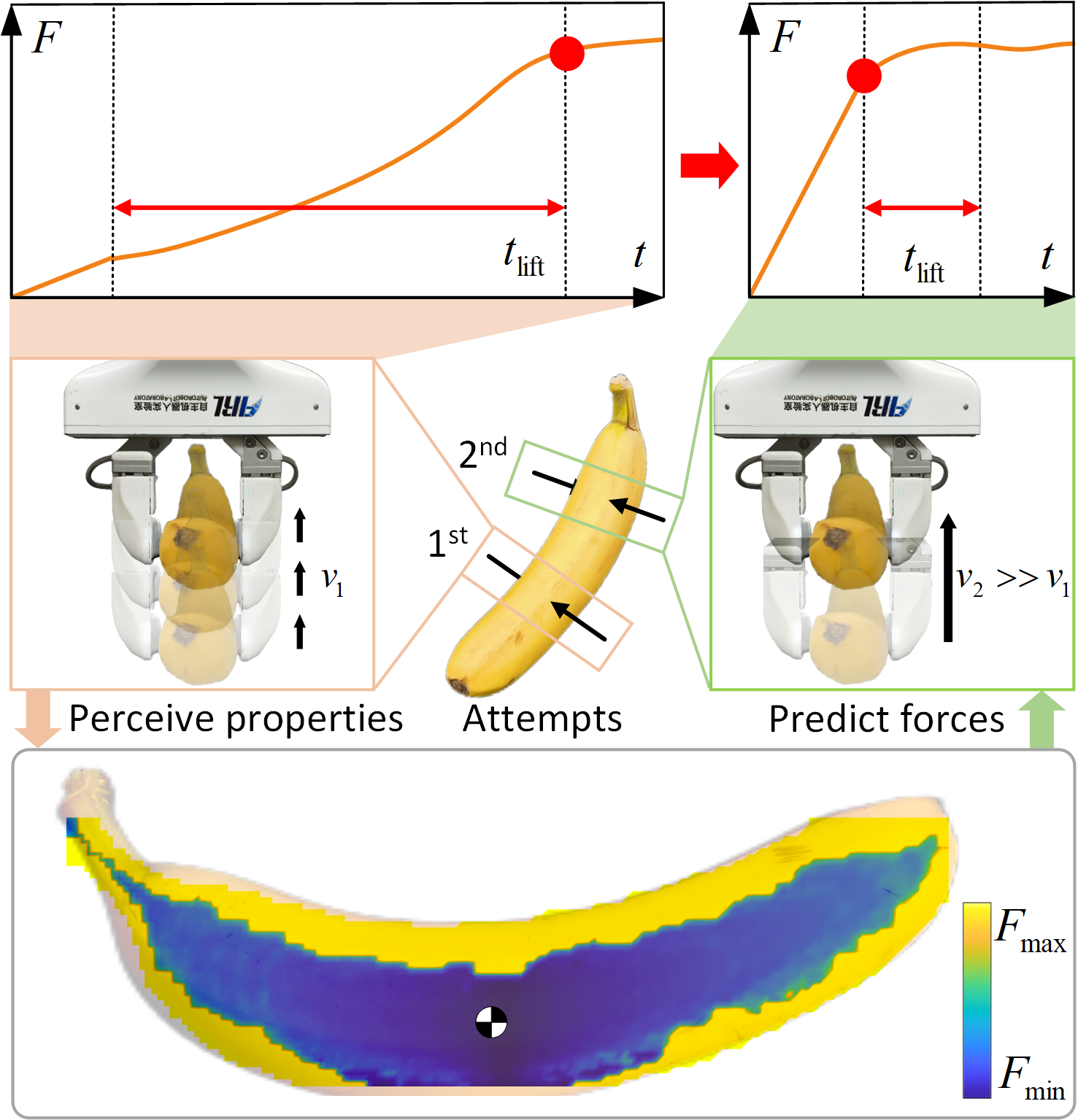}
    \caption{During the first successful grasping, the contact information is continuously and carefully monitored and the grasping force is adjusted accordingly. After slowly lifting the unknown object, its center of gravity is measured. In regrasping attempts, the required grasping force for different positions can be directly predicted before lifting the object. This approach eliminates the need for cautious lifting with continuous feedback, thereby improving grasping efficiency.}
    \label{fig::cover}
\end{figure} 

% 当已有的抓取无法胜任即将执行的任务时，就需要重新抓取。在重新抓取过程中，通过改变抓取位姿来满足任务需求。例如，在工具的重复使用过程中，可能需要抓取不同位置；在pick-and-place任务中，当初选的抓取位姿不佳时，需要改变位姿重新抓取；在同批次货品的搬运过程中，由于环境遮挡不同，需要以不同的位姿抓取物体。抓取位姿的改变将导致抓取力的变化。与此同时，物体的重力、形貌、摩擦系数等信息又未知。这两个特点使得再抓取的抓取力控制的难度显著提高。

When the current grasp is inadequate for the upcoming task, a regrasp is necessary. During the regrasping process, adjustments to the grasping pose are made to meet the task requirements. For instance, in the context of tool reuse, it may be necessary to grasp from different positions; in pick-and-place tasks, if the initially selected grasping pose is suboptimal, a change in pose is required for regrasping; and during the handling of a batch of goods, variations in environmental occlusions may necessitate grasping the objects from different poses. Changes in the grasping pose vary the grasping forces. At the same time, information regarding the object's weight, shape, and friction coefficient remains unknown. These two factors significantly increase the difficulty of controlling grasping forces during regrasping.

% While also relying on feedback-based methods \cite{prajith2021Grasping}, it is notable that humans actively leverage the feedback information to deepen their understanding of objects and predict the requirements for regrasping attempts attempts \cite{a.m.1993Memory}, \cite{evan2021Motor}. This anticipatory prediction process leads to notable improvements in grasping performance, including reduced decision time\cite{yasuhiro2011Developmental} and the ability to anticipate potential grasping outcomes\cite{moritz2020Making}.

\IEEEpubidadjcol

% 未知物体的首次抓取过程中，由于信息完全未知，只能通过反馈不断感知抓取信息，调整抓取力。在后续的多次抓取过程中，由于抓取姿态可能发生改变，抓取力需求也会对应改变，因此传统方法只能继续采用反馈方法调整抓取力。换言之，它将后续的同一未知物体的后续抓取尝试都当作首次抓取。在这些抓取过程中小心谨慎地根据实时反馈调整抓取力。由于是同一物体的多次抓取，即使抓取姿态发生改变，依旧存在某些感知到的重要信息。而传统方法忽视了这些重要信息，将抓取过多次的物体当作从来都没有抓取过，这种僵硬的谨慎可能严重限制了抓取效率。
Feedback-based force control methods have been widely applied in object grasping \cite{hao2014Stable}, \cite{Abi-Farraj2020}, particularly during the initial grasp of unknown objects. These methods rely on continuously perceiving real-time feedback to control the grasping force\cite{ziwei2022Review}. Regrasping attempts might include variations in grasping postures, thereby altering the force requirements. Consequently, conventional feedback-based methods cautiously adjust the grasping force, treating each regrasping attempt of the same unknown object as if it were the first. When consistently interacting with the same object, certain perceptible information remains unchanged or can be reused, even with variations in grasping postures. Unfortunately, these control strategies tend to overlook this valuable information, treating repeatedly grasped objects as if they had never encountered before. As a result, they adjust the grasping force solely based on current feedback during each attempt, which may limit overall grasping efficiency.

% 引入预测的方法的优点
Predictive methods, on the other hand, emphasize extracting essential, reusable insights from prior perception. This extracted information is used to predict subsequent grasping requirements, thereby reducing reliance on real-time and overly cautious feedback during grasping process. In particular, when the prediction of grasping force becomes sufficiently accurate, it becomes feasible to directly employ the predicted grasping force for successful grasping. Leveraging and integrating this predictive method into force control strategy holds the potential to significantly improve their grasping stability and efficiency. 

% 修改全文的抓取效率，力控效率只是手段，
Our objective is to improve grasping efficiency using the perception obtained in prior grasping attempts. We initially presented a method to improve efficiency by integrating prediction into the grasping process. Analyzing the general reach-lift process revealed the significant impact of the “lift” phase on grasping efficiency. We identified how the “lift” phase can be optimized through prediction. This optimization resulted in improved grasping efficiency after the first successful grasping. Subsequently, we developed a method for predicting grasping force that accounts for soft fingertip contact and asymmetric objects, validating the accuracy of these predictions through experimental testing. Ultimately, our comprehensive experiments verified the effectiveness and applicability of the method in enhancing grasping efficiency.

The contributions of this paper are:
\begin{itemize}
    \item [1)] Identifying the “lift” phase as the primary process affecting the grasping efficiency in the general reach-lift process, and outlining methods to optimize this phase;
    \item [2)] Proposing an efficient algorithm to determine the potential slip states and required grasping force considering the soft fingertip contact and asymmetric object;
    \item [3)] Developing a method to improve the efficiency of regrasping unknown objects without affecting the previous grasping process.
\end{itemize}

\section{Related Work}
\subsection{Grasp Force Control}
Force control is a crucial aspect of robot grasping. Inspired by human grasping perception and control strategies \cite{rolands.2009Coding}, current methods predominantly employ feedback-based control strategies. These approaches require real-time monitoring of stick surfaces \cite{ruomin2023Novel}, friction coefficient \cite{heba2021Realtime} or other relevant metrics \cite{James2021}, \cite{ruomin2022Incipient} to evaluate and adjust the grasping force.

Considering that one of the primary goals of grasping force control is to prevent object slippage, slip detection serves as a widely employed intermediate indicator for grasping force control \cite{Francomano2013}. Traditional methods for identifying slip states include neural network classification \cite{jianhua2018Slip} or incipient slip monitoring \cite{siyuan2019Maintaining}. A straightforward method to prevent slips is to increase the grasping force upon detecting proximity to slippage. However, because slippage and grasping failures often occur simultaneously, feedback-based methods typically require careful strategies and higher feedback frequencies to ensure successful grasping \cite{Veiga2018}. Moreover, slip detection requires real-time acquisition of extensive contact information, imposing high demands on sensor capabilities and feedback frequency \cite{Stachowsky2016}.

Prediction plays a pivotal role in human grasping strategies \cite{yasuhiro2011Developmental}. Humans' ability to anticipate the object properties before the grasping process begins allows for applying an appropriate amount of grasping force from the outset \cite{nikos2020Estimation}, thereby enhancing grasping efficiency. Furthermore, predicting the successful grasping force before slippage occurs can further improve the reliability of grasping, ensuring a higher success rate. 

Current research has employed predictive approaches using tactile information to predict the success rate of grasping before the actual grasp, demonstrating that utilizing tactile feedback can significantly increase the success rate of grasping various objects \cite{Calandra2017}. However, these studies primarily concentrate on the ultimate outcome of object grasping, such as whether it is lifted off the ground. They often overlook critical factors like rotational sliding between the object and fingers or variations in object postures. Additionally, many of these methods primarily consider rigid fingers, simplified object shapes and neglect the contact characteristics of soft fingertips. These factors collectively restrict the applicability of these prediction methods in various scenarios.

The core of controlling grasping force based on slippage prevention lies in ensuring that the grasping force surpasses the minimum successful grasping force required to prevent slippage. Building upon this concept, this paper proposes a method for modeling and solving the minimum successful grasping force applicable to soft fingers and asymmetric objects, along with outlining the basic sensor perception requirements. We use the minimum required grasping force obtained as a reference point to regulate the grasping force. Additionally, we use force prediction to improve grasping efficiency after the first grasping.

\subsection{Grasp Contact Modeling}
In the rigid contact model, the finger can exert normal and tangential forces on the object solely through the contact point. Employing the Coulomb friction model, the frictional cone typically delineates the boundaries of tangential forces, constrained by a limited normal force and a specified frictional coefficient \cite{jingyi20206DFC}, \cite{dean2021Novel}. 

Conversely, in the context of soft contact, the contact surface introduces a torsional moment about the contact normal. Here, the conventional frictional cone becomes inadequate, and the widely used approach shifts to the use of a limit surface. This surface illustrates the boundaries of the resultant tangential force and torsional torque \cite{amin2016Development}. The limit surface is constructed based on the power-law equation and the pressure distribution. The power law equation gives the relationship between the radius of the contact surface and the normal force for the soft fingers \cite{nicholas1999Modeling}. Combined with the pressure distribution within the contact surface, an approximate elliptical curve can be obtained, and each point on the curve represents the limit torque under different tangential forces under a given normal force. The versatility and intuitive nature of the limit surface method make it extensively adopted and refined for different soft materials \cite{amin2016Development}, \cite{amin2019Modeling}, \cite{Xu2021}. 

% 由\cite{Costanzo2020}等人开展的两指操作推动了极限面理论的应用，他们对极限面中与瞬心的概念进行了深入的分析与求解，并且仅通过两指就实现了平面物体的灵巧pivoting能力。在本文工作中，我们期望将极限面理论扩展到更加一般的非对称抓取场景中，即，两指接触平面非对称且带来两指非同时滑动的情景。
While the limit surface effectively reveals force-moment correlations on soft contact surfaces, it focuses on the contact analysis of individual fingers and fails to describe the overall finger-object contact system. The work by Costanzo et al. \cite{Costanzo2020} has advanced the application of the limit surface theory in two-finger manipulation. They conducted an in-depth analysis and resolution of the concepts related to the instantaneous center of rotation, successfully enabling force control for dexterous pivoting of planar objects using only two fingers. In this paper, we aim to extend the limit surface theory to more general asymmetric grasping scenarios, specifically where the two-finger contacts are asymmetric, resulting in non-simultaneous sliding of the fingers. Additionally, the limit surface theory introduces more unknowns and nonlinear relationships into the system, posing challenges in calculation convergence and efficiency. To mitigate these issues, we conducted modeling of the grasping system and optimized the solution process using mathematical transformations and physical approximations.

\section{Problem Statement}
% 本章节论述以下内容
% 说明研究目标是通过优化未知物体再抓取的力控制过程，提高再抓取的效率
% 首先，需要约束再抓取问题的研究对象不包括抓取位置选取，只是关注优化力控制过程。期望优化掉不断谨慎反馈对抓取效率的影响。具体方法层面，在第一次抓取未知物体时测量足够的物体信息，用于优化后续同一物体的再抓取过程。这一优化还需要考虑：再抓取过程中，抓取位置和物体姿态可能与第一次抓取不一样。
% ----definition
% 接着说明力控制过程的指标选为最小抓取力；确定成功抓取的定义；确定采用两指机械手完成抓取；说明采用软指机械手，并给出理论的基本假设。
%----framework
% 介绍框架。首先介绍required wrench，基于这一点将场景分割和抓取力求解分开。获取的方式决定了场景分割的方式，进一步使用就是抓取力求解的过程。
% 场景分割方面，将未知物体的多次抓取过程分为了第一次成功抓取和后续的再抓取过程。第一次成功抓取需要不断地谨慎感知反馈，并利用反馈来实时获取required wrench；后续再抓取过程可以利用第一次抓取的经验提前获取新颖物体的关键信息，如重力与重心位置等，利用这些经验来提前获取required wrench。
% 抓取力求解方面，输入信息为当前物体的required wrench，输出信息为此时物体所需的抓取力。上述框架有两个具体问题将在后续章节中解决：（1）再抓取时物体会与地面接触，仅通过重力相关信息无法获取每个时刻的物体所需抓取力；（2）两指软接触下，如何根据当前物体required wrench确定最小抓取力（并考虑单个手指发生滑动或者两个手指同时滑动）。

The research goal of this paper is to enhance the efficiency of regrasping by optimizing the force control process for unknown objects. For the purpose of this study, it is assumed that the gripper has already reached the required grasping position, allowing the focus to be on optimizing force control. The aim is to minimize the impact of continuous cautious feedback on grasping efficiency. The proposed method involves gathering sufficient object information during the first grasp of an unknown object to optimize the force control in subsequent regrasping attempts. The approach accounts for variations in the position and posture of the object during regrasping compared to the first grasp. Additionally, to enhance adaptability, the method must account for grasping scenarios with asymmetric contact, which may arise from factors such as object asymmetry or non-horizontal grasping postures.

\subsection{Definitions and Assumptions}

A fundamental requirement in grasping is maintaining the grasping force within the range bounded by the maximum safe grasping force (to avoid object damage) and the minimum force required for successful grasping (to prevent finger-object slippage). Considering the variable strength of different objects, a prudent approach is controlling the grasping force to exceed the minimum successful grasping force by a certain margin \cite{zherong2020Grasping}, \cite{He2020}. In light of these considerations, we establish the minimum required grasping force for preventing finger-object slippage as a reference point for control and optimization. Here grasping force is defined as the force applied by the motors to bring the fingers inward for grasping. 

Successful grasping is defined as the act of lifting an object from the ground while ensuring the object has not been crushed and there is no macro sliding between the fingers and the object. The parameters and variables are detailed in Table \ref{tab::parameters}. The parallel-fingers gripper holds widespread usage across industries, academia, and daily life due to its stability and structural simplicity \cite{Tai2016}, \cite{lionel2018Statistical}. As such, this study focuses on this particular gripper type.

\begin{table}[t]%调节图片位置，h：浮动；t：顶部；b:底部；p：当前位置
    \centering
    \caption{List of Notifications.}
    \label{tab::parameters}  
    \begin{tabular}{p{0.1\columnwidth}p{0.7\columnwidth}}%表格中的数据居中，c的个数为表格的列数	
        \toprule
        \multicolumn{2}{l}{Reference Frames} \\
        $\left\{\mathbb{G}\right\}$     & gripper reference frame\\
        $\left\{\mathbb{C}_i\right\}$     & reference frame of contact surface i\\
        \midrule
        \multicolumn{2}{l}{Parameters and Variables} \\
        $\boldsymbol{W}_\mathrm{req}$     & required wrench in $\left\{\mathbb{G}\right\}$\\
        $\boldsymbol{F}_\mathrm{req}$     & required force in $\left\{\mathbb{G}\right\}$\\
        $\boldsymbol{M}_\mathrm{req}$     & required moment in $\left\{\mathbb{G}\right\}$\\
        $\boldsymbol{F}_{\mathrm{ext}}$     & external force in $\left\{\mathbb{G}\right\}$\\
        $\boldsymbol{M}_{\mathrm{ext}}$     & external moment in $\left\{\mathbb{G}\right\}$\\
        $\widetilde{\boldsymbol{F}}_\mathrm{i}$     & feedback force of contact surface i 
                                                    in $\left\{\mathbb{C}_i\right\}$\\
        $\widetilde{\boldsymbol{M}}_\mathrm{i}$     & feedback moment of contact surface i 
                                                    in $\left\{\mathbb{C}_i\right\}$\\
        $\boldsymbol{r}_i$     & moment arms of contact surface i in $\left\{\mathbb{G}\right\}$\\
        $\boldsymbol{G}$     & object gravity in $\left\{\mathbb{G}\right\}$\\
        $\boldsymbol{l}$     & The position vector of the center of gravity of the object in  $\left\{\mathbb{G}\right\}$\\
        $\boldsymbol{F}_i$   & grasping force of finger i in $\left\{\mathbb{G}\right\}$\\
        $\boldsymbol{F}^{\mathrm{n}}_i$   & normal contact force of contact surface i in $\left\{\mathbb{G}\right\}$\\
        $\boldsymbol{F}^{\mathrm{t}}_i$   & tangential contact force of contact surface i in $\left\{\mathbb{G}\right\}$\\
        $\boldsymbol{F}^{\mathrm{t_1}}_i$   & projected tangential contact force of contact surface i in $\left\{\mathbb{G}\right\}$\\
        $\boldsymbol{F}^{\mathrm{t_2}}_i$   & projected tangential contact force of contact surface i in $\left\{\mathbb{G}\right\}$\\
        $\boldsymbol{e}^\mathrm{t_1}_i$   & normalized direction vector of $\boldsymbol{F}^{\mathrm{t_1}}_i$ in $\left\{\mathbb{G}\right\}$\\
        $\boldsymbol{e}^\mathrm{t_2}_i$   & normalized direction vector of $\boldsymbol{F}^{\mathrm{t_2}}_i$ in $\left\{\mathbb{G}\right\}$\\
        $\boldsymbol{e}^x$   & unit vector in the $x$-direction of $\left\{\mathbb{G}\right\}$\\
        $\boldsymbol{n}_i$   & normalized normal vector of contact surface $i$ in $\left\{\mathbb{G}\right\}$\\
        $\boldsymbol{M}_i$   & contact moment of contact surface $i$ in $\left\{\mathbb{G}\right\}$\\
        $\boldsymbol{M}^\mathrm{n}_i$   & normal contact moment of contact surface $i$ in $\left\{\mathbb{G}\right\}$\\
        $\mathcal{S} $   & slip state\\
        $\boldsymbol{G}_1$    & grasping matrix in one-finger slip state \\
        $\boldsymbol{G}_2$    & grasping matrix in two-finger slip state \\
        ${f}_1(\boldsymbol{X}_1)$    & optimization function in one-finger slip state \\
        $\boldsymbol{f}_2(\boldsymbol{X}_2)$   & optimization function in two-finger slip state \\
        $d_c$    & distance between center of contact surface to the center of rotation\\
        $R$    & radius of contact surface\\
        $\mu_i$    & friction coefficient of contact surface i\\
        \midrule
        \multicolumn{2}{l}{Material Parameters} \\
        $c$     &  contact coefficient of soft material \\
        $\gamma$    &exponent of the power-law equation \\
        $k$     &coefficient of pressure distribution \\
        \bottomrule
    \end{tabular}
\end{table}

When the grasping position or the external forces applied to the object change, significant torsional moments from the fingers might be required to prevent slippage between the object and the fingers. The torsional moment not only requires sufficient grasping force but also relies on a substantial contact surface between the fingers and the object. This objective can be addressed by employing soft fingertips. Therefore, we adopt soft fingertips, wherein the contact transitions from point contact to surface contact. To address the required force calculation with soft contacts, we make the following assumptions:

(1) The contact surface of the object is approximately flat. Even for objects with curved surfaces, the error introduced by this assumption is limited when the finger size is small or when the curvature of the contact surface is small;

(2) The friction coefficient within the contact surface remains consistent;

(3) The stiffness of the object is much higher than that of the soft finger.

\subsection{Framework}

\begin{figure*}[t]
  \centering
  \includegraphics[width = 1\linewidth]{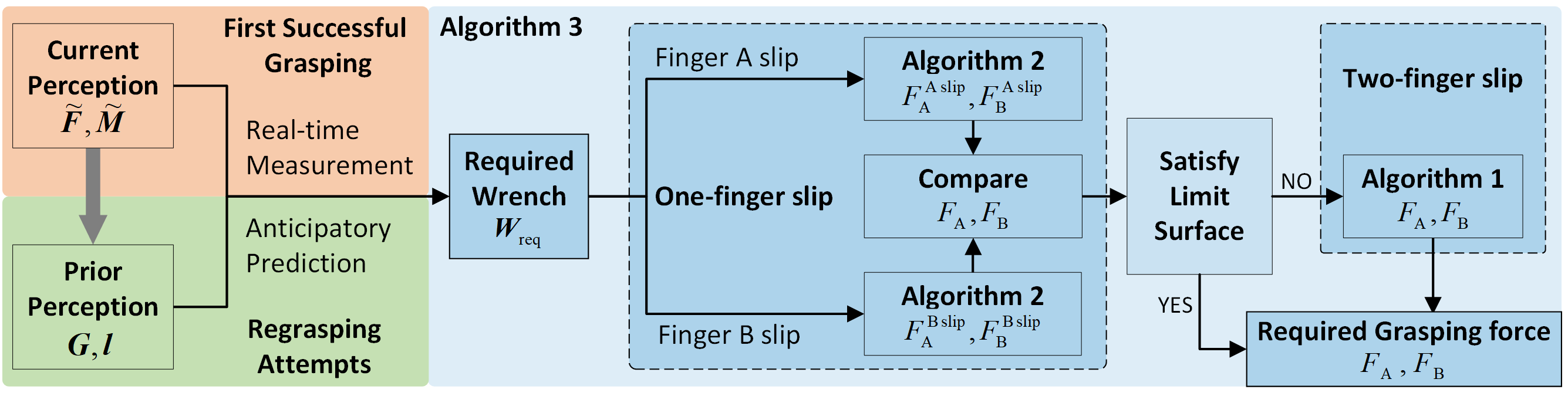}
  \caption{Our framework designed to determine the required grasping force in both the first successful grasping and the regrasping attempts. The required wrench can be derived from current perception during the first successful grasping or from prior perception during subsequent attempts. The required grasping force is then calculated using the required wrench, the friction coefficient, and the normal vector of the contact surface.}
  \label{fig::framework}
\end{figure*}

The key to achieving efficient grasping force control lies in determining the required grasping force for an object as quickly as possible, ideally even in advance. To address this, the paper proposes a grasping force solution framework, as shown in Fig. \ref{fig::framework}. The central variable in this framework is the required wrench, which encompasses both the resultant force and moment essential for the object to counteract external forces. Based on this variable, the process of determining the grasping force is divided into two specific tasks: the method for acquiring the required wrench in different scenarios and the algorithm for calculating the grasping force given a specific required wrench.

For scene segmentation, the multiple grasping processes of unknown objects are divided into the first successful grasp and the subsequent regrasping attempts. The first successful grasp requires continuous and careful feedback to determine the required wrench in real time. In contrast, the subsequent regrasping process can leverage the experience gained from the first grasp to obtain key information about the novel object in advance, such as its weight and center of gravity. This prior knowledge allows for the prediction of the required wrench, enhancing efficiency in regrasping. 

In the grasping force solution process, the input is the required wrench, and the output is the required grasping force. Given the two soft fingertips, it is crucial to identify the sliding state—whether one finger or both fingers are sliding—when the grasping force reaches the minimum required for the object. The grasping force is then determined based on this current state.

\section{Prediction Integration and Information Perception}
% 整个章节需要简明化

This section aims to integrate the prediction into the grasping process to improve its efficiency. An analysis of the typical reach-lift process is conducted. The primary process influencing grasping efficiency has been identified, along with methods to optimize it through prediction. Determining the grasping force requires acquiring the wrench required by the object. Consequently, methods for measuring or predicting the required wrench are also provided.

\subsection{Grasping Process Analysis}
% 说明抓取场景
Achieving a successful grasping takes three distinct steps: (1) Reach: where the gripper approaches the object surface and applies a predetermined grasping force without the need for continuous monitoring of intricate contact information; (2) lift: where the gripper raises the object until it is lifted off the ground. The minimum required grasping force continuously varies due to the interaction between the object and the ground. The primary goal of force control in this phase is to continuously measure the current contact information. This information is then used to adjust the current grasping force, ensuring it remains higher than the minimum required force to avoid slippage between the object and the gripper. (3) hold: the object detaches from the ground, and it experiences a constant gravitational force and torque.

\begin{figure}[t]
    \centering
    \includegraphics[width = 1\linewidth]{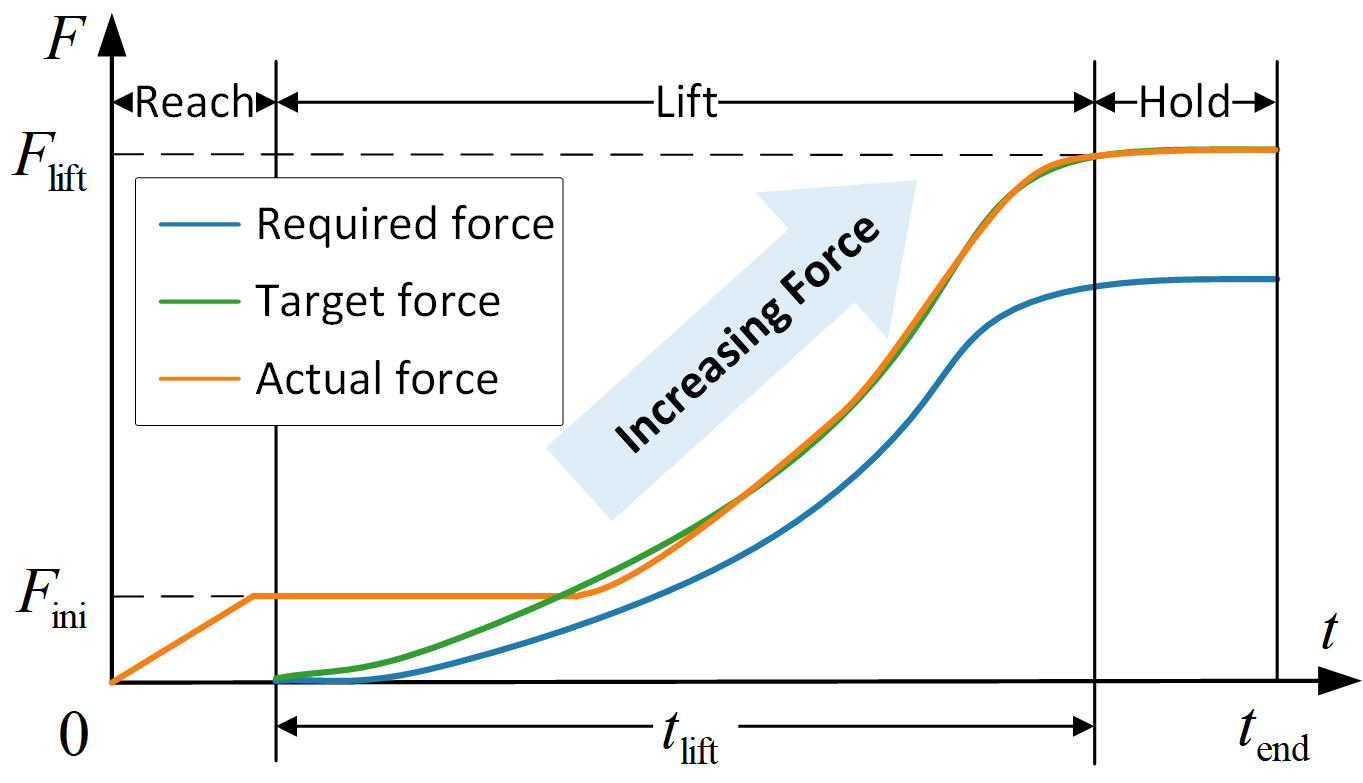}
    \caption{Relationship between grasping force and time $t$, divided into three steps.}
    \label{fig::F-t relationship}
\end{figure}

% 说明不同抓取力施加过程，只是描述过程力控的目的，特点
The application of grasping force can also be divided into three steps similar to the grasping process \cite{rolands.2009Coding}, as shown in Fig. \ref{fig::F-t relationship}. It is essential to note that setting the grasping force to an excessively high value to prevent slippage is not feasible, as it can potentially damage objects of unknown strength or fragility. The presence of delay and overshoot in the grasping force control system can lead to different fluctuations, causing deviations in the actual grasping force curve.

\subsection{Prediction Integration}
% 说明reach和hold阶段对抓取效率的影响很小
The three stages—“reach,” “hold,” and “lift”—affect overall grasping efficiency differently. In the “reach” phase, force control primarily adjusting to a pre-set initial grasping force. Well-developed algorithms allow for the quick application of specific grasping forces \cite{SUOMALAINEN2022104224}, resulting in minimal impact on efficiency during this phase. Similarly, in the “hold” phase, the object experiences only constant gravity and torque, leading to a stable required grasping force and reducing the need for continuous feedback, which also has a small effect on efficiency.

% 说明lift阶段对抓取效率的影响很大
In contrast, the “lift” phase significantly influences grasping efficiency, as the required grasping force continuously changes due to varying ground support. The target force for control varies moment to moment, and the need for feedback complicates the process, leading to delays in sensor perception. These factors collectively result in a scenario where the gripper can only cautiously and slowly lift the object. Otherwise, the actual force does not match the required force, resulting in potential slippage.

% 说明不能直接预测lift阶段的抓取力
Although predicting grasping force during the “lift” phase could enhance efficiency by reducing the need for continuous feedback, it is challenging due to unpredictable factors like ground conditions, grasping postures, and ground friction. Thus, optimizing the “lift” phase may require indirect methods.

% 说明抓取过程特点，提出效率提高方法
Notably, during the “lift” phase, as ground support diminishes, the required grasping force increases, peaking when the object transitions into the “hold” phase. This peak value represents the minimum required grasping force for successful lifting. Ensuring the grasping force exceeds this value can enhance grasping success. By leveraging the predictability of the “hold” phase, this peak value can be anticipated and set as the initial grasping force, bypassing limitations imposed by cautious force control during the “lift” phase. Furthermore, in the “hold” phase, predicting grasping force can eliminate the need to estimate complex object-environment interactions.

\subsection{Required Wrench Estimation}

% 介绍所需力旋量的基本定义。
Estimating the required wrench for an object to counteract external forces is essential, whether using current feedback or prior perception. This estimation is used for predicting the minimum required grasping force. As illustrated in Fig. \ref{fig::framework}, current feedback is employed to estimate the required wrench during the first grasp, while prior perception is used for regrasping attempts. 

% 基于反馈的方法，说明和所需力旋量和直接的反馈量之间的差别。
The required wrench does not simply equate to the current force and torque measured by sensors. For example, when grasping a cuboid, each finger may apply substantial normal grasping forces, but gravity acting vertically downward may not necessitate any horizontal resultant force. Consequently, summing the forces and torques from each finger can result in a calculated grasping force that is significantly higher than what the object actually needs, posing risks such as object crushing.

\begin{figure}[t]
    \centering
    \includegraphics[width = 0.7\linewidth]{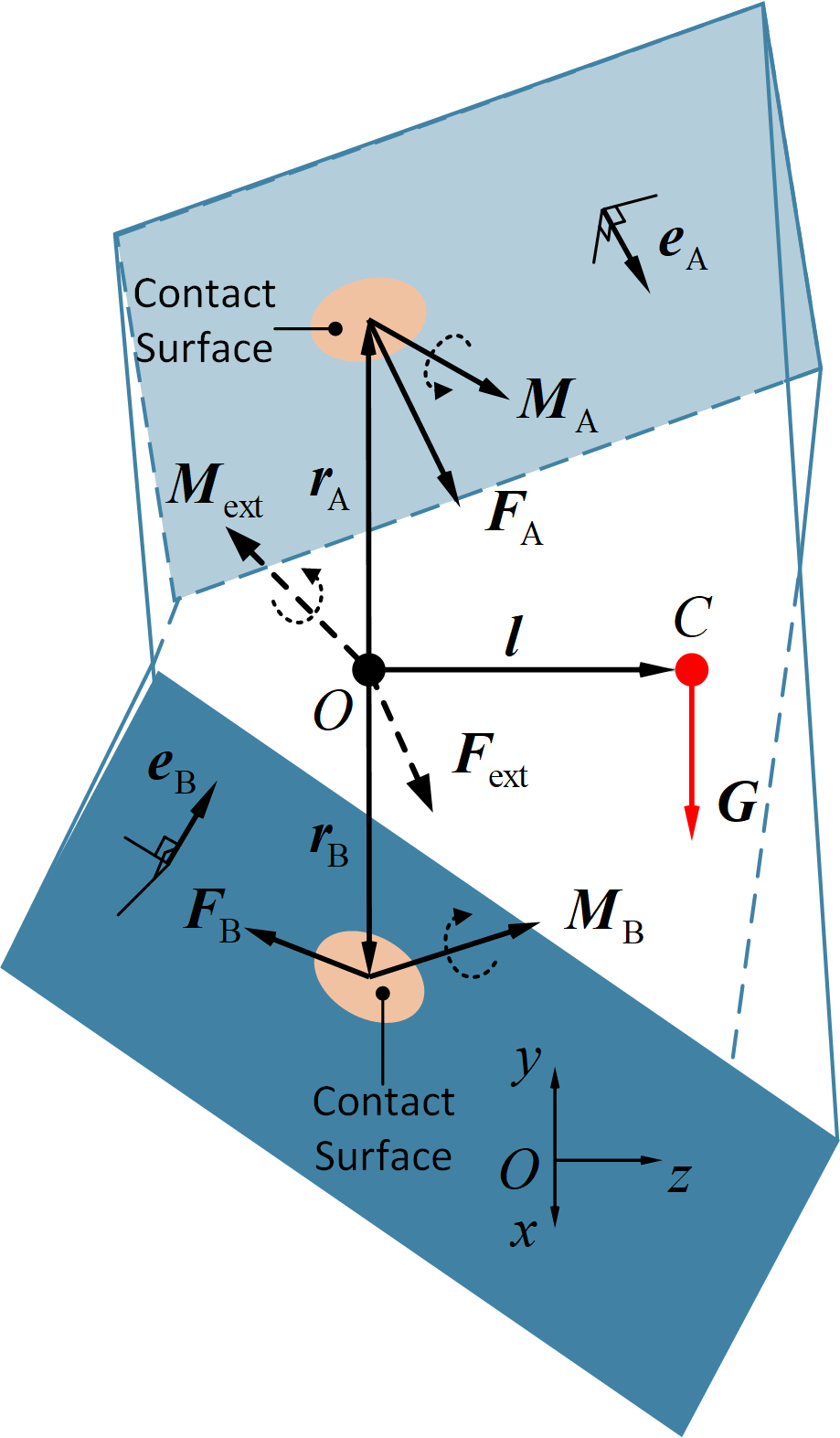}
    \caption{Force analysis depicting contact force and other external forces acting on the object.}
    \label{fig::force analyze}
\end{figure}

% 说明如何基于反馈的方法如何计算。
To address these issues, determining the required wrench for the object must involve a force balance approach. Specifically, the object requires a wrench from the fingers to counter external loads, including gravity. In other words, assessing the required wrench is equivalent to evaluating the external forces acting on the object at that moment:
\begin{equation}
    \label{equ::required wrench}
    \boldsymbol{W}_\mathrm{req}=\left[\boldsymbol{F}_\mathrm{req}, \boldsymbol{M}_{\mathrm{req}}\right],
\end{equation}
\begin{equation}
    \label{equ::relationship between external force and required force}
    \boldsymbol{F}_{\mathrm{req}}=- \boldsymbol{F}_{\mathrm{ext}}, \boldsymbol{M}_{\mathrm{req}}=- \boldsymbol{M}_{\mathrm{ext}},
\end{equation}
where $\boldsymbol{F}_\mathrm{ext}$ and $\boldsymbol{M}_{\mathrm{ext}}$ denote the external force and moment, respectively. $\boldsymbol{F}_\mathrm{req}$, $\boldsymbol{M}_{\mathrm{req}}$, and $\boldsymbol{W}_\mathrm{req}$ denote the required force, moment, and wrench.
Since the object is solely influenced by the fingers and the external environment, including gravity, we can directly calculate the external force and moment using sensor feedback and the principle of force and moment balance:
\begin{equation}
  \label{equ::require force from feedback}
  \boldsymbol{F}_{\mathrm{ext}}=-\left(\widetilde{\boldsymbol{F}}_\mathrm{A}+ \widetilde{\boldsymbol{F}}_\mathrm{B}\right),
\end{equation}
\begin{equation}
  \label{equ::require torque from feedback}
  \boldsymbol{M}_{\mathrm{ext}}=-\left(\widetilde{\boldsymbol{M}}_\mathrm{A}+\widetilde{\boldsymbol{M}}_\mathrm{B} +\boldsymbol{r}_\mathrm{A}\times \widetilde{\boldsymbol{F}}_\mathrm{A} + \boldsymbol{r}_\mathrm{B}\times \widetilde{\boldsymbol{F}}_\mathrm{B}\right),
\end{equation}
% 说明坐标系建立问题，说明基本参数定义
where $\widetilde{\boldsymbol{F}}_\mathrm{A}$, $\widetilde{\boldsymbol{F}}_\mathrm{B}$, $\widetilde{\boldsymbol{M}}_\mathrm{A}$ and $\widetilde{\boldsymbol{M}}_\mathrm{B}$ denote the forces and torques acquired from the sensor feedback, with the contact surface center as the origin. These measurements are expressed in the gripper reference frame $\left\{\mathbb{G}\right\}$. The coordinate axes of the reference frames of two contact surfaces $\left\{\mathbb{C}_i\right\}$ are aligned parallel to the coordinate axes of the gripper reference frame $\left\{\mathbb{G}\right\}$, hence the rotation matrix $\boldsymbol{R}=\mathbf{I}$. As a result, the forces and moments represented in both coordinate systems are numerically identical. For better clarity in subsequent equations, forces in both coordinate systems are denoted using the same expressions. The same principle applies to moments. $\boldsymbol{r}_\mathrm{A}$ and $\boldsymbol{r}_\mathrm{B}$ denote the moment arms, measured from the gripper coordinate system as the origin to the endpoints at the contact surface centers, as illustrated in Fig. \ref{fig::force analyze}. By substituting equations (\ref{equ::require force from feedback}) and (\ref{equ::require torque from feedback}) into equations (\ref{equ::required wrench}) and (\ref{equ::relationship between external force and required force}), we derive the required torque for the regrasping attempts.

% 说明基于预测的方法如何计算
In the regrasping attempts, besides the wrench applied by the fingers, an object is solely affected by gravity. When the center of mass position and the gravitational force magnitude are known, the external wrench exerted on the object can be directly computed using:
\begin{equation}
    \label{equ::require force from prediction}
    \boldsymbol{F}_{\mathrm{ext}}=\boldsymbol{G},
\end{equation}
\begin{equation}
    \label{equ::require torque from prediction}
    \boldsymbol{M}_{\mathrm{ext}}=\boldsymbol{l}\times\boldsymbol{G},
\end{equation}
% 说明基本参数定义
where $\boldsymbol{G}$ denotes the gravity and $\boldsymbol{l}$ denotes the position vector of the center of gravity of the object in the gripper reference frame $\left\{\mathbb{G}\right\}$. Substituting equations (\ref{equ::require force from prediction}) and (\ref{equ::require torque from prediction}) into equations (\ref{equ::required wrench}) and (\ref{equ::relationship between external force and required force}) yields the required torque for the regrasping attempts.

Consider that once the center of mass and the magnitude of gravity are identified, it becomes feasible to compute the required wrench for the object across various grasping positions and orientations, given the information of the relative posture between the gripper and the object. This implies that for estimating the required wrench, there is no need to grasp the object at every possible position to accumulate experience about it. Simply measuring the center of mass and gravity magnitude is adequate for estimating the wrench even for positions or orientations that have never been previously grasped.

\subsection{Gravity Measurement}
% 介绍如何获取重心
To predict the required wrench using equation (\ref{equ::require force from prediction}) and equation (\ref{equ::require torque from prediction}), it is essential to measure the gravity of the object. When the object is not in contact with the ground, it experiences forces solely from the fingers and gravity. The torque sensed by the finger sensors and the torque induced by gravity cancel each other out. In essence, we can directly estimate the gravitational force using the feedback values from the current sensors. By combining equations (\ref{equ::require force from feedback}) and (\ref{equ::require force from prediction}), the gravity can be measured:
\begin{equation}
    \label{equ::measurement of G}
    \boldsymbol{G}=-\left(\widetilde{\boldsymbol{F}}_\mathrm{A}+\widetilde{\boldsymbol{F}}_\mathrm{B}\right).
\end{equation}

Similarly, by combining equations (\ref{equ::require torque from feedback}) and (\ref{equ::require torque from prediction}), we can obtain:
\begin{equation}
    \label{equ::measurement of Gl}
    \boldsymbol{l}\times\boldsymbol{G}=-\left(\widetilde{\boldsymbol{M}}_\mathrm{A}+\widetilde{\boldsymbol{M}}_\mathrm{B} +\boldsymbol{r}_\mathrm{A}\times \widetilde{\boldsymbol{F}}_\mathrm{A}+\boldsymbol{r}_\mathrm{B}\times \widetilde{\boldsymbol{F}}_\mathrm{B}\right).
\end{equation}

However, due to the irreversibility property of cross product, the vector $\boldsymbol{l}$ cannot be directly solved for. Only the general solution of $\boldsymbol{l}$ can be obtained:
\begin{equation}
    \label{equ::general solution of l}
    \boldsymbol{l}=\boldsymbol{l}_0+\gamma \boldsymbol{G}, \, \gamma\in \mathbf{R},
\end{equation}
\begin{equation}
    \label{equ::special solution of l}
    \boldsymbol{l}_0=\frac{|\widetilde{\boldsymbol{M}}_\mathrm{r}|}{|\boldsymbol{G}|} \frac{\boldsymbol{G}\times \widetilde{\boldsymbol{M}}_\mathrm{r}}{|\boldsymbol{G}\times \widetilde{\boldsymbol{M}}_\mathrm{r}|},
\end{equation}
% 说明基本参数定义
where $\widetilde{\boldsymbol{M}}_\mathrm{r} = -\left(\widetilde{\boldsymbol{M}}_\mathrm{A}+\widetilde{\boldsymbol{M}}_\mathrm{B} +\boldsymbol{r}_\mathrm{A}\times \widetilde{\boldsymbol{F}}_\mathrm{A}+\boldsymbol{r}_\mathrm{B}\times \widetilde{\boldsymbol{F}}_\mathrm{B}\right)$ denotes the resultant moment due to the contact forces between the finger and the object. The variable $\gamma$ can take any real value, meaning that the center of gravity can move arbitrarily along the line of gravity.

Equations (\ref{equ::general solution of l}) and (\ref{equ::special solution of l}) show that each instance of sensor feedback establishes a line indicating the location of the center of gravity. When two or more non-parallel lines are obtained from separate feedback instances, their intersection can pinpoint the center of gravity. This process does not require multiple grasping attempts; instead, utilizing feedback information twice during a single grasp—by adjusting finger orientation, for example—can effectively determine the center of mass.

% 说明结合了抓取过程，不用特意设计重心求解
Integrating essential grasping information, particularly gravity and its position, into the grasping process eliminates the need for complex measurement methods or additional platforms. By leveraging the measured gravity and center of mass position, we can predict the required wrench in advance, as illustrated in Fig. \ref{fig::grasping scenarios}.

\section{Gripping Force Solution}
% 本章节确定给定物体外力旋量下的最小抓取力求解方法。由于考虑了软接触手指以及非对称抓取状态，仅通过平衡方程以及极限面约束（见附录）是远不足以高效求解当前抓取力的。具体而言，非对称抓取导致两指不一定都会发生滑动，需要判断滑移的情况，并不同情况补充相应的方程。同时，软接触手指将施加给物体高维的力旋量以及极限面理论带来的非线性项将增加系统求解难度。上述问题的解决方法将在本章节具体阐述。

This chapter establishes a method for determining the minimum grasping force under a given external wrench acting on an object. Considering the soft-contact fingers and asymmetric grasping states, relying solely on equilibrium equations and limit surface constraints (see Appendix) is insufficient for efficiently solving the minimum grasping force. Specifically, the asymmetric grasp may result in non-uniform sliding of the two fingers, necessitating an assessment of the sliding conditions and the incorporation of corresponding equations for each scenario. Furthermore, the soft-contact fingers apply high-dimensional wrenches to the object, and the nonlinearities introduced by limit surface theory complicate the system's resolution. The solutions to these challenges will be detailed in this chapter.

\subsection{Gripping Contact Modeling}

\begin{figure}[t]
    \centering
    \includegraphics[width = 1\linewidth]{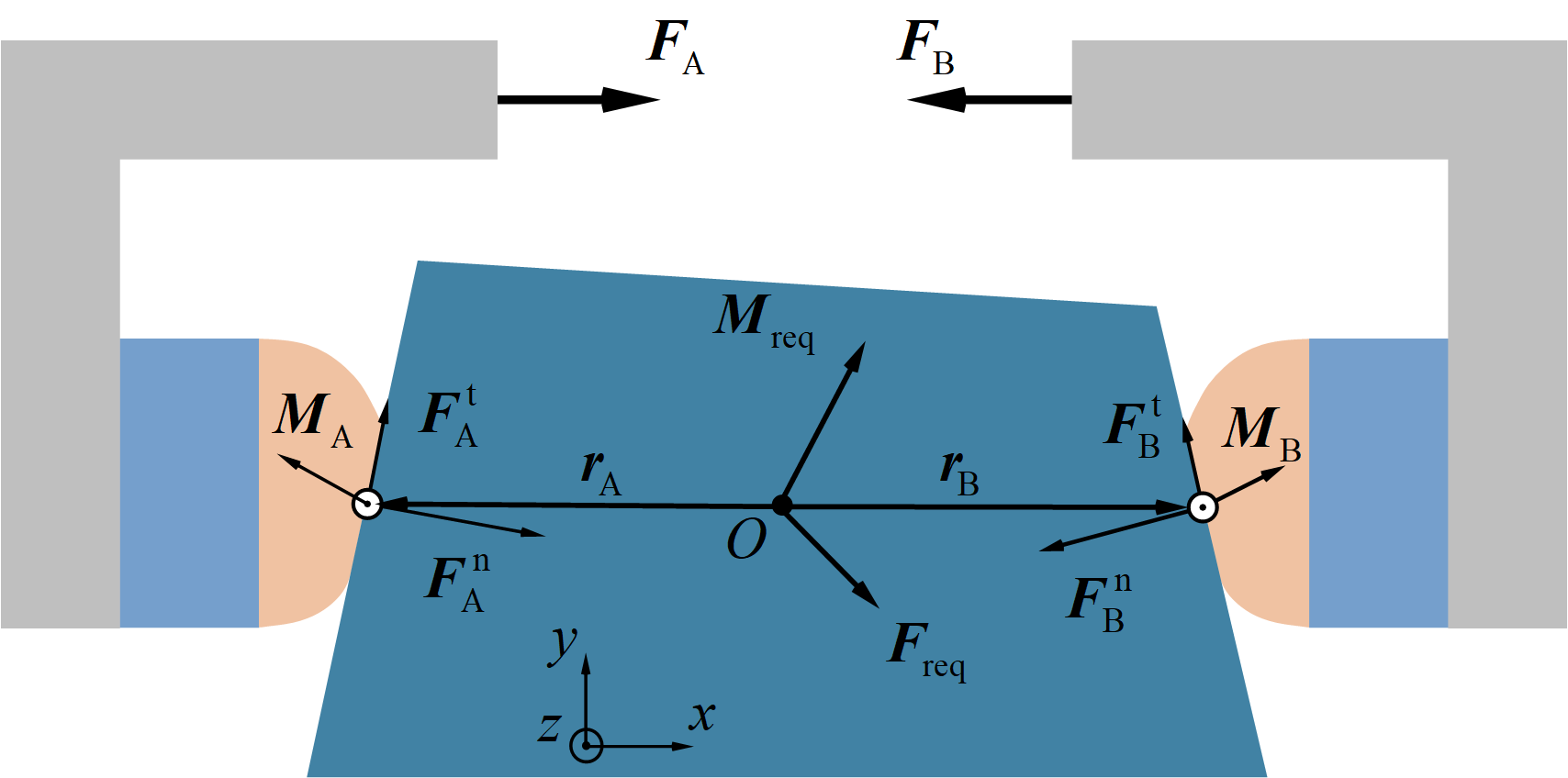}
    \caption{Force analysis depicting the grasping force and the force and moment required by the object.}
    \label{fig::force relation}
\end{figure}

Inappropriate grasping force will cause the object to slip (under a small force) or crush (above a large force). As the fragility of the object is unknown, a safe approach is to maintain the grasping force above the minimum required force by a certain margin. In other words, given a required wrench $\boldsymbol{W}_\mathrm{req} = [\boldsymbol{F}_\mathrm{req}, \boldsymbol{M}_\mathrm{req}]$, we need to find the minimum required grasping force $\boldsymbol{F}_\mathrm{A}$ and $\boldsymbol{F}_\mathrm{B}$, as shown in Fig. \ref{fig::force relation}.

\begin{figure}[t]
    \centering
    \includegraphics[width = 0.8\linewidth]{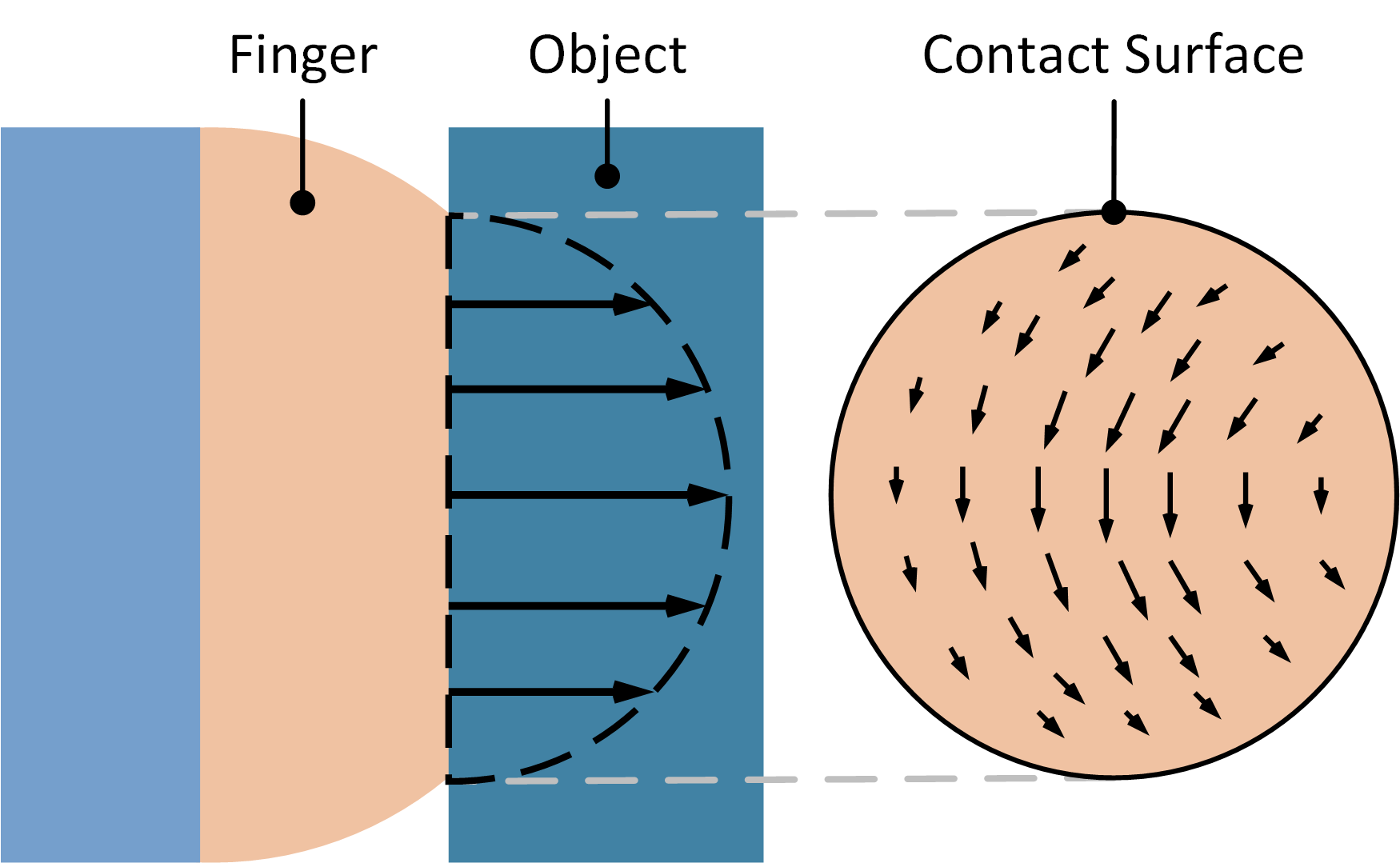}
    \caption{Soft contact model.}
    \label{fig::soft model}
\end{figure}

% 说明软指的特点
In the contact surface, the normal forces and frictional forces exhibit variability from one point to another, as illustrated in Fig. \ref{fig::soft model}. The direction of the frictional forces depends on the relative motion between the points and the object. Consequently, calculating the maximum tangential force cannot be achieved by merely multiplying the friction coefficient with the normal force. Additionally, under soft contact conditions, distributed forces not only supply the contact force but also generate the contact moment.

When deformation is not negligible, a soft contact model should be considered, where the contact surface is no longer a single point. The forces distributed within this surface can be represented as resultant force and moment, with their points of application situated at the center of the contact surface. To analyze the forces, we project the resultant force into two components: the normal force $\boldsymbol{F}^\mathrm{n}$ and the tangential force $\boldsymbol{F}^\mathrm{t}$. As shown in Fig. \ref{fig::force relation}, the grasping forces $\boldsymbol{F}_i$ and the projected contact forces $\boldsymbol{F}_i^\mathrm{n}$, $\boldsymbol{F}_i^\mathrm{t}$ satisfy the following equation:
\begin{equation}
    \label{equ::force jacobian}
    \boldsymbol{F}_i = \boldsymbol{J}_i\left(\boldsymbol{F}_i^\mathrm{n} + \boldsymbol{F}_i^\mathrm{t}\right),
\end{equation}
where $i=\mathrm{A,B}$ denotes the contact surface $\mathrm{A}$ or $\mathrm{B}$. The matrices $\boldsymbol{J}_i$ denotes the force jacobian in contact surface $i$. For a two-fingers parallel gripper, the force jacobians are $\boldsymbol{J}_\mathrm{A} = \left[1, 0, 0\right]$ and $\boldsymbol{J}_\mathrm{B} = \left[-1, 0, 0\right]$. 

We align the $x$-axis parallel of the gripper reference frame $\left\{\mathbb{G}\right\}$ to the direction of the grasping force. The tangential forces $\boldsymbol{F}^{t}_i$ in the contact surface $\mathrm{A}$ and $\mathrm{B}$ can be calculated as follows:
\begin{equation}
    \label{equ::projected tangential force}
    \boldsymbol{F}_i^\mathrm{t} = \boldsymbol{F}_i^\mathrm{t_1} + \boldsymbol{F}_i^\mathrm{t_2},
\end{equation} 
\begin{equation}
    \label{equ::direction of Ft1}
    \boldsymbol{F}_i^\mathrm{t_1} = \boldsymbol{F}_i \cdot \boldsymbol{e}^\mathrm{t_1}_i=F_i^\mathrm{t_1} \left({\boldsymbol{n}_i\times\boldsymbol{e}^x}\right)/{\left|\boldsymbol{n}_i\times\boldsymbol{e}^x\right|},
\end{equation} 
\begin{equation}
    \label{equ::direction of Ft2}
    \boldsymbol{F}_i^\mathrm{t_2} = \boldsymbol{F}_i \cdot \boldsymbol{e}_i^\mathrm{t_2}=F_i^\mathrm{t_2} \left(\boldsymbol{e}_i^\mathrm{t_1}\times\boldsymbol{n}_i\right),
\end{equation}
where the tangential force $\boldsymbol{F}_i^\mathrm{t_1}$ is parallel to the $x$-$z$ plane, while $\boldsymbol{F}_i^\mathrm{t_2}$ is parallel to the contact surface $i$ and perpendicular to $\boldsymbol{F}_i^\mathrm{t_2}$. The vector ${n}_i$ denotes the normalized normal vector of contact surface $i$. The vectors $\boldsymbol{e}_i^\mathrm{t1}$ and $\boldsymbol{e}_i^\mathrm{t_2}$ denote the normalized direction vector of $\boldsymbol{F}_i^\mathrm{t1}$ and $\boldsymbol{F}_i^\mathrm{t_2}$, respectively. The vector $\boldsymbol{e}^x$ denotes the unit vector in the $x$-direction of gripper reference frame $\left\{\mathbb{G}\right\}$.

% 介绍状态方程（受力平衡方程），说明现有的六个方程无法求解接触面合力
To ensure the balance of forces and moment on the object, it is necessary to satisfy:
\begin{equation}
    \label{equ::force balance}
    \boldsymbol{F}_\mathrm{A}^\mathrm{n} +\boldsymbol{F}_\mathrm{A}^\mathrm{t} +\boldsymbol{F}_\mathrm{B}^\mathrm{n} +\boldsymbol{F}_\mathrm{B}^\mathrm{t} =\boldsymbol{F}_\mathrm{req},
\end{equation}
\begin{equation}
    \label{equ::moment balance}
    \begin{split}
    \boldsymbol{M}_\mathrm{A}+\boldsymbol{M}_\mathrm{B} +\boldsymbol{r}_\mathrm{A}\times\left(\boldsymbol{F}^\mathrm{n}_\mathrm{A}+\boldsymbol{F}^\mathrm{t}_\mathrm{A}\right) \\
    +\boldsymbol{r}_\mathrm{B}\times\left(\boldsymbol{F}^\mathrm{n}_\mathrm{B}+\boldsymbol{F}^\mathrm{t}_\mathrm{B}\right)=\boldsymbol{M}_{\mathrm{req}}.
    \end{split}
\end{equation}

As shown in Fig. \ref{fig::force relation}, the forces $\boldsymbol{F}^\mathrm{n}_i$ and $\boldsymbol{F}^\mathrm{t}_i$ denote the normal and tangential forces at the contact surface $i$. The moments $\boldsymbol{M}_i$ represent the contact moment of contact surface $i$. 

% 说明无法求解的问题，说明问题的本质原因
The above equations (\ref{equ::force jacobian})--(\ref{equ::moment balance}) alone do not suffice to solve for the grasping force. Firstly, when minimizing the grasping force, it becomes essential to ascertain the sliding states of the two fingers, i.e., whether both fingers are sliding or if only one finger is sliding. Secondly, the correlation between the torque and force provided by the contact surface when the fingers are prone to sliding has not yet been established.

\subsection{Two-finger Slip State}

% 先分析两指同时滑动，说明需要引入极限面理论，增加接触面的切向力-切向力矩关系，此时可以用于求解双指同时滑动
When a substantial torsional force is required to be exerted by the fingers, it often leads to simultaneous slippage occurring on both fingers. When both fingers slip simultaneously, the resultant normal moment $M_i^\mathrm{n}$ and the resultant tangential force $F_i^\mathrm{t}$ satisfy the limit surface constraint, as depicted in the Appendix.
\begin{equation}
    \label{equ::relation between force and moment}
    {\left(F_i^\mathrm{t}\right)^{2}}/{\left(F_{i}^\mathrm{t,\max}\right)^{2}}+{\left(M_i^\mathrm{n}\right)^{2}}/{\left(M_{i}^\mathrm{n,\max}\right)^{2}}=1,
\end{equation}
\begin{equation*}
    F_{i}^\mathrm{t,\max} = \mu_i F_{i}^\mathrm{n}, M_i^\mathrm{n} = M_i^\mathrm{n, \max}|_{d_{\mathrm{c}}=0},     
\end{equation*}
where $\mu_i$ is the friction coefficients of contact surface $i$, and $d_{\mathrm{c}, i}$ is the distance between the center of rotation to the center of the contact surface.

The torque on the contact surface arises from two types of distributed forces: normal forces and tangential friction forces. As indicated by the pressure distribution (\ref{equ::pressure distribution}), the normal force is symmetrically distributed, leading to a net moment of zero. Thus, the torque generated at the contact surface is primarily due to friction. Since the distributed friction forces act tangentially on the contact surface, they can only generate torque around the contact normal. Previous research suggests that while the assumption of symmetrical normal force distribution may not always hold, the resulting error is negligible \cite{amin2016Development}. Therefore, it can be assumed that the resultant moment from the fingers aligns closely with the normal to the contact surface:
\begin{equation}
    \label{equ::moment approximation}
    \| \boldsymbol{M}_i - \boldsymbol{M}^\mathrm{n}_i \| \approx 0.
\end{equation}

% 说明双指滑动时方程特性，引入线性化方程组+非线性优化方法
Since the number of unknowns matches the number of the equation, the minimal grasping force can be solved by using (\ref{equ::force jacobian})--(\ref{equ::moment approximation}). However, (\ref{equ::moment balance}) and (\ref{equ::relation between force and moment}) are nonlinear, presenting computational efficiency and convergence challenges. To overcome this, we utilized specialized mathematical techniques and optimized the solving process.

We linearized (\ref{equ::moment balance}) by choosing point $O$ in Fig. \ref{fig::force analyze} as the rotational center and used coordinate axes $x$, $y$, and $z$ as the pivots. We then substituted (\ref{equ::projected tangential force})--(\ref{equ::direction of Ft2}), (\ref{equ::moment approximation}) into (\ref{equ::moment balance}) and utilize the properties of the scalar triple product$\left(\boldsymbol{A} \times \boldsymbol{B}\right) \cdot \boldsymbol{C} = \left(\boldsymbol{B} \times \boldsymbol{C}\right) \cdot \boldsymbol{A}$, the scalar forms of the moment balance (\ref{equ::moment balance}) are

\begin{align}
    \label{equ::projected moment direction x} M_{\mathrm{req}}^x = & M_\mathrm{A} n_\mathrm{A}^x + M_\mathrm{B} n_\mathrm{B}^x, \\
    M_{\mathrm{req}}^{y} = & -F_\mathrm{A}^\mathrm{n} r_\mathrm{A} n_\mathrm{A}^{z}-F_\mathrm{A}^\mathrm{t1} r_\mathrm{A} e_\mathrm{A}^\mathrm{t1,z}-F_\mathrm{A} ^\mathrm{t_2} r_\mathrm{A} e_\mathrm{A}^\mathrm{t_2,z} \notag \\
    \label{equ::projected moment direction y} &+F_\mathrm{B}^\mathrm{n} r_\mathrm{B} n_\mathrm{B}^{z}+F_\mathrm{B}^\mathrm{t1} r_\mathrm{B} e_\mathrm{B}^\mathrm{t1,z}+F_\mathrm{B}^\mathrm{t_2} r_\mathrm{B} e_\mathrm{B}^\mathrm{t_2,z}\\
    &+M_\mathrm{A} n_\mathrm{A}^{y}+M_\mathrm{B} n_\mathrm{B}^{y}, \notag \\
    M_{\mathrm{req}}^{z} = & +F_\mathrm{A}^\mathrm{n} r_\mathrm{A} n_\mathrm{A}^{y}+F_\mathrm{A}^\mathrm{t1} r_\mathrm{A} e_\mathrm{A}^\mathrm{t1,y}+F_\mathrm{A}^\mathrm{t_2} r_\mathrm{A} e_\mathrm{A}^\mathrm{t_2,y} \notag \\
    \label{equ::projected moment direction z} &-F_\mathrm{B}^\mathrm{n} r_\mathrm{B} n_\mathrm{B}^{y}-F_\mathrm{B}^\mathrm{t1} r_\mathrm{B} e_\mathrm{B}^\mathrm{t1,y} - F_\mathrm{B}^\mathrm{t_2} r_\mathrm{B} e_\mathrm{B}^\mathrm{t_2,y}\\
    & + M_\mathrm{A} n_\mathrm{A}^{z}+M_\mathrm{B} n_\mathrm{B}^{z}. \notag     
\end{align}

By substituting the (\ref{equ::projected tangential force})--(\ref{equ::direction of Ft2}) into (\ref{equ::force balance}), and combine (\ref{equ::projected moment direction x})--(\ref{equ::projected moment direction z}), an underdetermined system of the equation is obtained:
\begin{equation}
    \label{equ::underdetermined system 2}
    \boldsymbol{G}_2 \boldsymbol{X}_2 = \boldsymbol{W}_{\mathrm{req}},
\end{equation}
\iffalse
\begin{equation}
    \label{equ::matrix M}
    \boldsymbol{G}_2 = \begin{bmatrix}
            n_\mathrm{A}^\mathrm{x} & e_\mathrm{A}^\mathrm{t1,x} & e_\mathrm{A}^\mathrm{t_2,x} & n_\mathrm{B}^\mathrm{x} & e_\mathrm{B}^\mathrm{t1,x} & e_\mathrm{B}^\mathrm{t_2,x} & 0 & 0\\
            n_\mathrm{A}^\mathrm{y} & e_\mathrm{A}^\mathrm{t1,y} & e_\mathrm{A}^\mathrm{t_2,y} & n_\mathrm{B}^\mathrm{y} & e_\mathrm{B}^\mathrm{t1,y} & e_\mathrm{B}^\mathrm{t_2,y} & 0 & 0\\
            n_\mathrm{A}^\mathrm{z} & e_\mathrm{A}^\mathrm{t1,z} & e_\mathrm{A}^\mathrm{t_2,z} & n_\mathrm{B}^\mathrm{z} & e_\mathrm{B}^\mathrm{t1,z} & e_\mathrm{B}^\mathrm{t_2,z} & 0 & 0\\
            0 & 0 & 0 & 0 & 0 & 0 & e_\mathrm{A}^x & e_\mathrm{B}^x\\
            -r_\mathrm{A} n_\mathrm{A}^\mathrm{z} & -r_\mathrm{A} e_\mathrm{A}^\mathrm{t1,z} & -r_\mathrm{A} e_\mathrm{A}^\mathrm{t_2,z} & r_\mathrm{B} n_\mathrm{B}^\mathrm{z} & r_\mathrm{B} e_\mathrm{B}^\mathrm{t1,z} & r_\mathrm{B} e_\mathrm{B}^\mathrm{t_2,z} & e_\mathrm{A}^y & e_\mathrm{B}^y\\
            r_\mathrm{A} n_\mathrm{A}^\mathrm{y} & r_\mathrm{A} e_\mathrm{A}^\mathrm{t1,y} & r_\mathrm{A} e_\mathrm{A}^\mathrm{t_2,y} & -r_\mathrm{B} n_\mathrm{B}^\mathrm{y} & -r_\mathrm{B} e_\mathrm{B}^\mathrm{t1,y} & -r_\mathrm{B} e_\mathrm{B}^\mathrm{t_2,y} & e_\mathrm{A}^z & e_\mathrm{B}^z\\
            \end{bmatrix}
\end{equation}
\fi
where $\boldsymbol{X}_2=\left[F_\mathrm{A}^\mathrm{n},F_\mathrm{A}^\mathrm{t1},F_\mathrm{A}^\mathrm{t_2},F_\mathrm{B}^\mathrm{n},F_\mathrm{B}^\mathrm{t1},F_\mathrm{B}^\mathrm{t_2},M_\mathrm{A},M_\mathrm{B}\right]^{\mathrm{T}}$ denotes the unknown contact force and moment to be solved. The matrix $\boldsymbol{G}_2\in \mathbf{R}^{6\times 8}$ denotes the grasping matrix of two-finger slip state. The vector $\boldsymbol{W}_{\mathrm{req}}=\left[F_{\mathrm{req}}^{x},F_{\mathrm{req}}^{y},F_{\mathrm{req}}^{z},M_{\mathrm{req}}^{x},M_{\mathrm{req}}^{y},M_{\mathrm{req}}^{z}\right]^{\mathrm{T}}$ denotes the required wrench. The solution of an underdetermined system of equations is
\begin{equation}
    \label{equ::underdetermined system 2: solution}
    \boldsymbol{X}_2 = \boldsymbol{p}_2 + \boldsymbol{Z}_2 \boldsymbol{q}_2, \,
    \boldsymbol{p}_2=\boldsymbol{G}_2^{\mathrm{T}}\left(\boldsymbol{G}_2\boldsymbol{G}_2^{\mathrm{T}}\right)^{-1} \boldsymbol{W}_{\mathrm{req}}.
\end{equation}

The special solution $\boldsymbol{p}_2$ is the minimal norm least square solution of equation (\ref{equ::underdetermined system 2}). Columns of $\boldsymbol{Z}_2$ are the basis vectors in the nullspace of the grasping matrix $\boldsymbol{G}_2$. The general solution $\boldsymbol{q}_2\in \mathbf{R}^{2\times 1}$ is an unknown vector. By substituting equations (\ref{equ::underdetermined system 2: solution}) to the limit surface (\ref{equ::relation between force and moment}), a relatively simple nonlinear system of equations is obtained:
\begin{equation}
    \label{equ::underdetermined system 2: optimization functions}
    \begin{aligned}
        \boldsymbol{f}_2(\boldsymbol{X}_2) &= {\left(F_i^\mathrm{t}(\boldsymbol{X}_2)\right)^{2}}/{\left(F_{i}^\mathrm{t,\max}(\boldsymbol{X}_2)\right)^{2}}\\
        &+ {\left(M_i^\mathrm{n}(\boldsymbol{X}_2)\right)^{2}}/{\left(M_{i}^\mathrm{n,\max}(\boldsymbol{X}_2)\right)^{2}}-1.
    \end{aligned}
\end{equation}

Subsequently, nonlinear optimization is employed to determine the current grasping force. We utilized the classical Levenberg-Marquardt optimization method \cite{10.1007/BFb0067700} to solve for the nonlinear system. To address potential issues related to multiple solutions in nonlinear systems, we initiated the search from various starting points, conducting multiple rounds of optimizations. The obtained solutions were then substituted into equation (\ref{equ::force jacobian}) to calculate the current grasping force. Among the multiple solutions, we selected the smallest real number solution as the final result. The procedural algorithm is presented in Algorithm (\ref{alg::calculation of two finger slip}).

\IncMargin{1em}
\begin{algorithm}[t]
	\KwData{Required wrench $\boldsymbol{W}_{\mathrm{req}}$, friction coefficient $\mu_i$, Normal vector of contact surface $\boldsymbol{n}_i$}
	\KwResult{Required force ${F}_{\mathrm{A}}, {F}_{\mathrm{B}}$}
    \BlankLine 

    initialization\;
    $\boldsymbol{q}_{\mathrm{ini}} = \left\{[1;1],[1;-1],[-1;1],[-1;-1]\right\}$\;
    $\boldsymbol{G}_2 \gets \mu_i,\boldsymbol{n}_i$\;   %// Equation (\ref{equ::projected tangential force})-(\ref{equ::direction of Ft2}), (\ref{equ::force balance}), and (\ref{equ::projected moment direction x})-(\ref{equ::projected moment direction z})\;
    $\boldsymbol{p}_2,\boldsymbol{Z}_2 \gets \boldsymbol{G}_2,\boldsymbol{W}_{\mathrm{req}}$\;% // Equation (\ref{equ::underdetermined system 2}), (\ref{equ::underdetermined system 2: solution})\;
    \For{$k \gets 1$ \KwTo $4$}{
        $\boldsymbol{q}_2 = \boldsymbol{q}_{\mathrm{ini}}(k)$\;
        // Levenberg-Marquardt optimization
        
        min $\boldsymbol{f}_2(\boldsymbol{X}_2)$ subject to $\boldsymbol{X}_2 = \boldsymbol{p}_2 + \boldsymbol{Z}_2 \boldsymbol{q}_2$\;
        $\boldsymbol{X}_2^{(k)} = \boldsymbol{X}_2|_{\min \boldsymbol{f}_2(\boldsymbol{X}_2)}$\;
        $F_\mathrm{A}^{(k)},F_\mathrm{B}^{(k)} \gets \boldsymbol{X}_2^{(k)}$\;
    }
    $F_{\mathrm{A}},F_{\mathrm{B}} \gets \mathop{\min}\limits_{k}{(F_\mathrm{A}^{(k)},F_\mathrm{B}^{(k)})}$\;
    \caption{Solving Force for Two-finger Slip State.}
    \label{alg::calculation of two finger slip}
\end{algorithm}
\DecMargin{1em} 

\subsection{One-finger Slip State}
\begin{figure}[t]
    \centering
    \includegraphics[width = 0.4\linewidth]{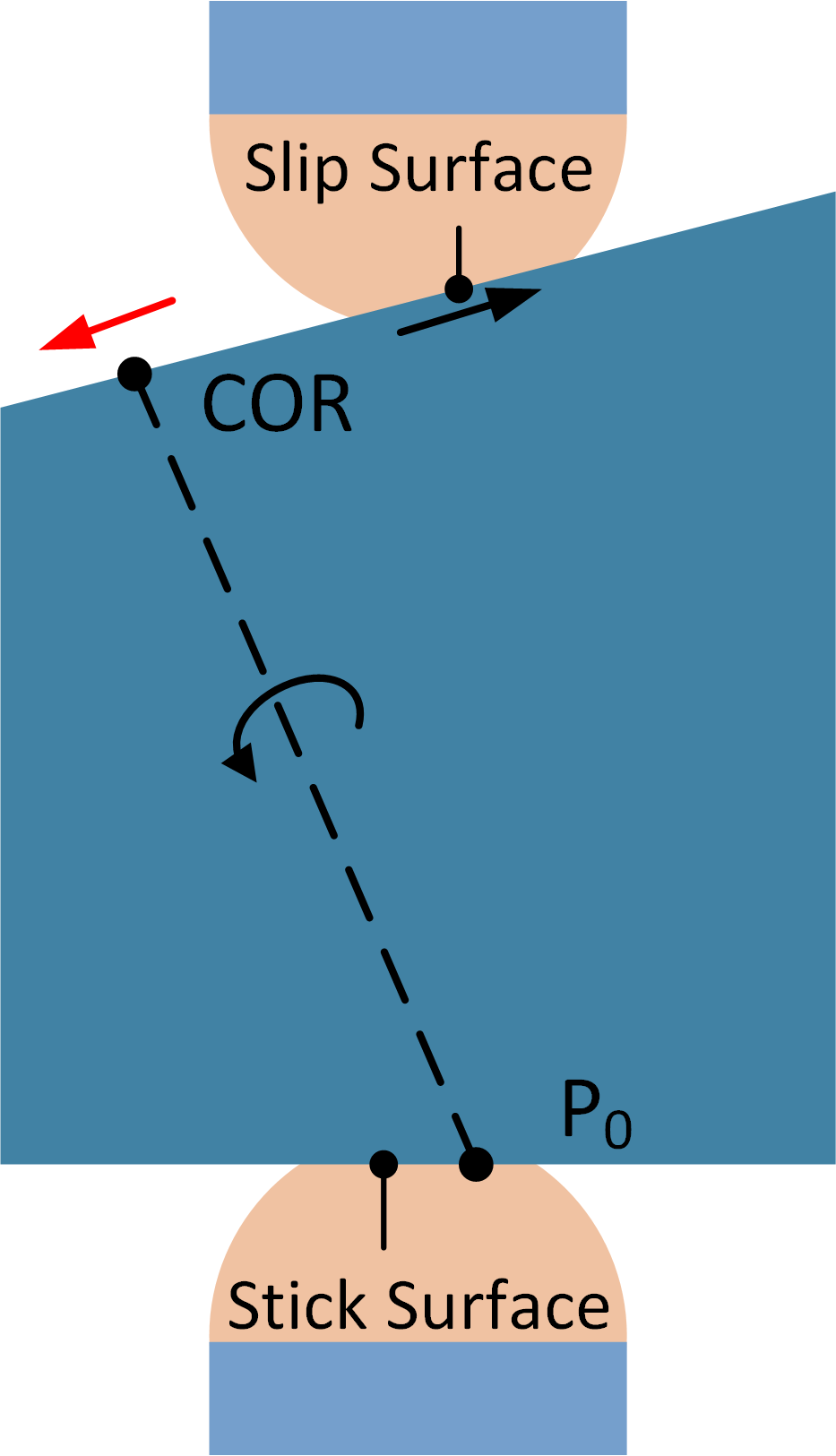}
    \caption{Kinematic analysis of one-finger slip state.}
    \label{fig::one finger slip}
\end{figure}

% 说明即使引入了极限面理论，也无法求解单指发生滑动的问题。
In practical grasping scenarios, there are instances where only one finger undergoes slippage. This situation is particularly noticeable when there exists a disparity in the friction coefficient or when one finger is positioned above the object and the other below, as shown in Fig. \ref{fig::one finger slip}. When only a single finger slips, the contact surface of the stick finger no longer satisfies the limit surface theory equation (\ref{equ::relation between force and moment}), rendering Algorithm (\ref{alg::calculation of two finger slip}) inapplicable.

% 说明单指滑动时，瞬心必须远离接触面中心。
During one-finger slippage, an instantaneous center of rotation (COR) exists for the sliding finger (see Appendix). As the object adheres to the surface, its movement can be seen as rotation around an axis connecting the COR and a point ($\mathrm{P}\mathrm{0}$) on the stick surface. The displacement of contact points increases with their distance from the rotation axis. With relative sliding between the finger and the object, points on the contact surface lag behind, resulting in slippage. If the COR is far from the contact surface, the relative displacement between the finger and contact points increases, leading to more sliding. On the other hand, as points approach $\mathrm{P}\mathrm{0}$, the fingers can match the object's motion, preventing slippage. 

As the COR moves closer to the contact surface, the displacements in the slip surface decrease, aligning more closely with those in the stick surface. When the COR is close enough to the contact surface, or even enters the contact surface, the object displacement within the slip surface may even be smaller than that within the stick contact surface. This unexpected reversal implies that the stick contact surface, contrary to expectations, becomes more susceptible to sliding compared to its actual sliding state. Therefore, for the one-finger sliding state, the COR of the slip surface should not be too close to the contact surface; otherwise, the stick surface will also experience sliding.

\begin{figure}[t]
    \centering
    \includegraphics[width = 1\linewidth]{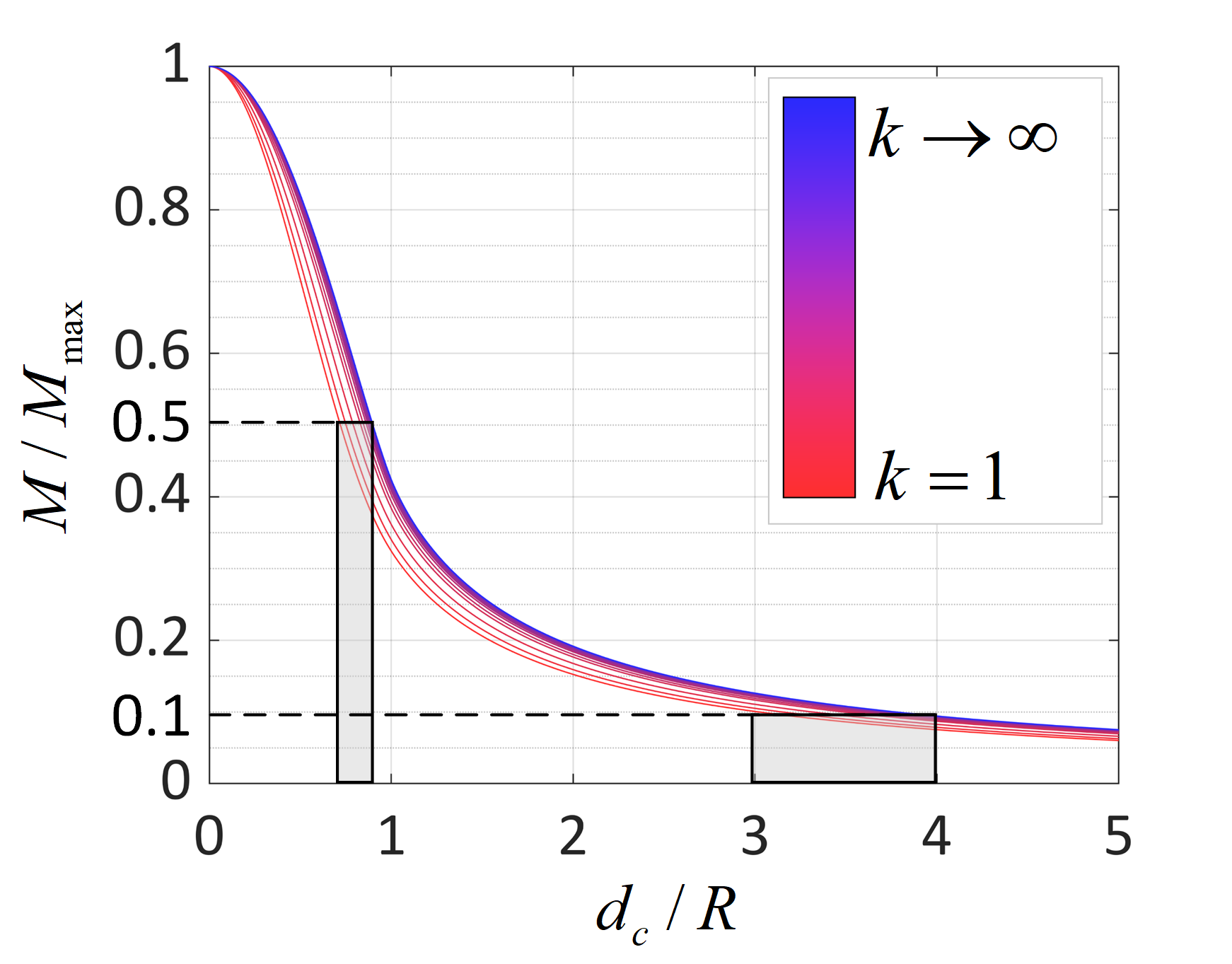}
    \caption{The relationship between the torque exerted by the contact surface and the position of the COR for different pressure distribution coefficients $k$.}
    \label{fig::torque graph}
\end{figure}

% 说明瞬心距和力矩的关系,给出极限面理论中的力与力矩曲线
The direction of friction forces at the contact surface is influenced by the COR. When the COR is distant, the friction directions align closely, and if located at infinity, they are parallel. With a limited normal force, this alignment reduces the torque generated by the contact surface. As shown in Fig. \ref{fig::torque graph}, as the distance ratio $d_c/R$ increases, the torque decreases rapidly. When the COR remains within the contact surface ($d_\mathrm{c}/R>1$), the torque can drop to half its maximum, and when $d_\mathrm{c}/R>4$, it can fall below 0.1 times its maximum value.

% 最后给出方法总结，补充说明我们用一种代价很小的方式解决这个问题（避免仿真等方法）。
Our observations indicate that as $d_c/R$ increases, the torque provided by the contact surface diminishes quickly. In light of the COR of the sliding finger moving away from the contact surface during the one-finger slip state, we adopt an approximation. This approximation assumes that the torque transmitted by the finger through the slip surface is negligible, resembling a rigid contact model. This simplified approach avoids complex material mechanics analyses and time-consuming finite element simulations. Ultimately, our solution addresses the challenge of determining grasping force during the one-finger slip state in an approximate yet effective manner:
\begin{equation}
    \label{equ::torque approximation}
    \boldsymbol{M}_\mathrm{slip}^\mathrm{n} \approx  0,
\end{equation}
where $\boldsymbol{M}_\mathrm{slip}^\mathrm{n}$ denotes the normal moment of sliding contact surface. For slip surface, the normal force $\boldsymbol{F}_\mathrm{slip}^\mathrm{n}$ and tangential force $\boldsymbol{F}_\mathrm{slip}^\mathrm{t}$ need to satisfy:
\begin{equation}
    \label{equ::force relationship without moment}
    \boldsymbol{F}^\mathrm{n}_\mathrm{slip} = \mu_\mathrm{slip} \boldsymbol{F}^\mathrm{t}_\mathrm{slip}.
\end{equation}

We have also employed a linearization method similar to the method of the two-finger slip state. Substitute the (\ref{equ::projected tangential force})--(\ref{equ::direction of Ft2}) into (\ref{equ::force balance}), and combine (\ref{equ::projected moment direction x})--(\ref{equ::projected moment direction z}), (\ref{equ::torque approximation}), we obtain another underdetermined system of the equation: 
\begin{equation}
    \label{equ::underdetermined system 1}
    \boldsymbol{G}_1 \boldsymbol{X}_1 = \boldsymbol{W}_{\mathrm{req}},
\end{equation}
where $\boldsymbol{X}_1=\left[F_\mathrm{A}^\mathrm{n},F_\mathrm{A}^\mathrm{t1},F_\mathrm{A}^\mathrm{t_2},F_\mathrm{B}^\mathrm{n},F_\mathrm{B}^\mathrm{t1},F_\mathrm{B}^\mathrm{t_2},M_\mathrm{stick}\right]^{\mathrm{T}}$ denotes the unknowns to be solved. $M_\mathrm{stick}$ denotes the moment offered by the stick surface. The matrix $\boldsymbol{G}_1\in \mathbf{R}^{6\times 7}$ denotes the grasping matrix of one-finger slip state. The vector $\boldsymbol{W}_{\mathrm{req}}=\left[F_{\mathrm{req}}^{x},F_{\mathrm{req}}^{y},F_{\mathrm{req}}^{z},M_{\mathrm{req}}^{x},M_{\mathrm{req}}^{y},M_{\mathrm{req}}^{z}\right]^{\mathrm{T}}$ denotes the required wrench. The solution of an underdetermined system of equations is
\begin{equation}
    \label{equ::underdetermined system 1: solution}
    \boldsymbol{X}_1 = \boldsymbol{p}_1 + \boldsymbol{Z}_1 q_1, \,
    \boldsymbol{p}_1=\boldsymbol{G}_1^{\mathrm{T}}\left(\boldsymbol{G}_1\boldsymbol{G}_1^{\mathrm{T}}\right)^{-1} \boldsymbol{W}_{\mathrm{req}}.
\end{equation}

The special solution $\boldsymbol{p}_1$ is the minimal norm least square solution of equation (\ref{equ::underdetermined system 1}). Columns of $\boldsymbol{Z}_1$ are the basis vectors in the nullspace of the grasping matrix $\boldsymbol{G}_1$. The general solution $q_1\in \mathrm{R}$ is the unknown to be solved. By substituting equations (\ref{equ::underdetermined system 1: solution}) to (\ref{equ::force relationship without moment}) and (\ref{equ::projected tangential force}), we obtain a quadratic equation:
\begin{equation}
    \label{equ::underdetermined system 1: optimization function}
    \begin{aligned}
        f_1(\boldsymbol{X}_1) &= (F^\mathrm{n}_\mathrm{slip}(\boldsymbol{X}_1))^2 - \mu_\mathrm{slip}^2 (F^\mathrm{t}_\mathrm{slip}(\boldsymbol{X}_1))^2.
    \end{aligned}
\end{equation}

The quadratic equation can be directly solved without the need for nonlinear optimization methods. The obtained solutions were then substituted into equation (\ref{equ::force jacobian}) to calculate the current grasping force. The smallest real number solution among the two solutions was then chosen as the obtained result of the solution. The procedural algorithm is presented in Algorithm (\ref{alg::calculation of one finger slip}).

\IncMargin{1em}
\begin{algorithm}[t]
	\KwData{Required wrench $\boldsymbol{W}_{\mathrm{req}}$, friction coefficient $\mu_i$, normal vector of contact surface $\boldsymbol{n}_i, $slip state $\mathcal{S}$}
	\KwResult{required grasping force ${F}_{\mathrm{slip}}, {F}_{\mathrm{stick}}$, contact force and moment of the stick contact surface ${F}_\mathrm{stick}^\mathrm{n}, {F}_\mathrm{stick}^\mathrm{t}, {M}_\mathrm{stick}^\mathrm{n}$}
    \BlankLine 

    initialization\;
    $\boldsymbol{G}_1 \gets \mu_i,\boldsymbol{n}_i,\boldsymbol{M}_\mathrm{slip}^\mathrm{n},\mathcal{S}$\;
    $\boldsymbol{p}_1,\boldsymbol{Z}_1 \gets \boldsymbol{G}_1,\boldsymbol{W}_{\mathrm{req}}$\;
    $\boldsymbol{X}_1 = \boldsymbol{p}_1 + \boldsymbol{Z}_1 \boldsymbol{q}_1$\;
    // Solve for the quadratic equation $f_1(\boldsymbol{X}_1)$

    $\boldsymbol{X}_1^{(1)}, \boldsymbol{X}_1^{(2)} \gets f_1(\boldsymbol{X}_1) = 0$\;
    $F_i^{(1)}, F_i^{(2)} \gets \boldsymbol{X}_1^{(1)}, \boldsymbol{X}_1^{(2)}$\;
    $k \gets \min (F_i^{(1)},F_i^{(2)})$\;
    ${F}_{\mathrm{slip}}, {F}_{\mathrm{stick}},{F}_\mathrm{stick}^\mathrm{n}, {F}_\mathrm{stick}^\mathrm{t}, {M}_\mathrm{stick}^\mathrm{n} \gets k,(\boldsymbol{X}_1^{(1)}, \boldsymbol{X}_1^{(2)})$\;

    \caption{Solving Force for One-finger Slip State.}
    \label{alg::calculation of one finger slip}
\end{algorithm}
\DecMargin{1em} 

\subsection{Gripping Force Solution}
% 根据上述内容，给出最小可行抓取力计算方法与流程。包括单指-双指滑动判据等。
Based on the Algorithms (\ref{alg::calculation of two finger slip}) and (\ref{alg::calculation of one finger slip}), the minimum required grasping force for both sliding states can be determined, resulting in two distinct sets of solutions. However, only one type of sliding state occurs at the point where the grasping force reaches its minimum value. Therefore, it is necessary to determine which sliding state will occur. 

In the case of the one-finger slip state, only the mechanical characteristics of the slip surface were analyzed, as represented by equation (\ref{equ::torque approximation}). In practice, however, the stick surface must also adhere to specific conditions to maintain its adherence. Precisely, to sustain the stick surface, the moments and tangential forces exerted by the contact surface must not surpass the limit surface. The sliding state will transition from the one-finger slip state to the two-finger slip state if the following conditions are satisfied:
\begin{equation}
    \label{equ::condition for trasition}
    {\left(F^\mathrm{t}_\mathrm{stick}\right)^{2}}/{\left(F^\mathrm{t,\max}_\mathrm{stick}\right)^{2}}+{\left(M^\mathrm{n}_\mathrm{stick}\right)^{2}}/{\left(M^\mathrm{n,\max}_\mathrm{stick}\right)^{2}}>1,
\end{equation}
\begin{equation*}
    F_\mathrm{stick}^\mathrm{t,\max} = \mu_\mathrm{stick} F_\mathrm{stick}^\mathrm{n}, M_\mathrm{stick}^\mathrm{n,\max} = M^\mathrm{n}|_{d_{\mathrm{c}}=0},     
\end{equation*}
where $F^\mathrm{t}_\mathrm{stick}$ and $M^\mathrm{n}_\mathrm{stick}$ denote the tangential force and normal moment of the stick surface, respectively. $F^\mathrm{t,\max}_\mathrm{stick}$ and $M^\mathrm{n,\max}_\mathrm{stick}$ denote the maximum tangential force and normal moment that the stick surface can offer.

We begin by assuming a one-finger slip state and employ Algorithm (\ref{alg::calculation of one finger slip}) to compute the grasping force along with the forces and moments facilitated by the stick surface. Subsequently, an evaluation is conducted to determine whether the stick surface complies with equation (\ref{equ::condition for trasition}). Satisfying this equation signifies the transition to a two-finger slip state. Algorithm (\ref{alg::calculation of two finger slip}) is then utilized to determine the current grasping force. 

When dealing with a situation where only one finger slips, it is important to determine which finger slide first. This determination method requires separately computing the minimum required grasping force for the scenarios where either finger A or B undergoes sliding. If the calculated grasping force for finger A sliding surpasses that of finger B sliding, it indicates that finger A initiates the sliding action, and vice versa. This approach is employed because, as the grasping force gradually decreases, it first reaches the larger value among the solutions. At this point, the corresponding finger will start to slip, leading to the grasping failure. The overall computational procedure is outlined in Algorithm (\ref{alg::calculation of force}).

\IncMargin{1em}
\begin{algorithm}[t]
	\KwData{Required wrench $\boldsymbol{W}_{\mathrm{req}}$, friction coefficient $\mu_i$, Normal vector of contact surface $\boldsymbol{n}_i$}
	\KwResult{required grasping force ${F}_{\mathrm{A}}, {F}_{\mathrm{B}}$, slip state $\mathcal{S}$}
    \BlankLine 

    initialization\;
    $\left\{{F}_{\mathrm{slip}}, {F}_{\mathrm{stick}},{F}_\mathrm{stick}^\mathrm{n}, {F}_\mathrm{stick}^\mathrm{t}, {M}_\mathrm{stick}^\mathrm{n}\right\}|_\mathrm{A\,slip} \gets $ Algorithm\_2 $(\boldsymbol{W}_{\mathrm{req}}, \mu_i, \boldsymbol{n}_i, \mathcal{S} = \mathrm{A\, slip})$\;
    $\left\{{F}_{\mathrm{slip}}, {F}_{\mathrm{stick}},{F}_\mathrm{stick}^\mathrm{n}, {F}_\mathrm{stick}^\mathrm{t}, {M}_\mathrm{stick}^\mathrm{n}\right\}|_\mathrm{B\,slip} \gets $ Algorithm\_2 $(\boldsymbol{W}_{\mathrm{req}}, \mu_i, \boldsymbol{n}_i, \mathcal{S} = \mathrm{B\, slip})$\;
    \eIf{$({F}_{\mathrm{slip}}+{F}_{\mathrm{stick}})|_\mathrm{A\,slip} > ({F}_{\mathrm{slip}}+{F}_{\mathrm{stick}})|_\mathrm{B\,slip}$}{
        $\mathcal{S} =$ A slip\;
        ${F}_{\mathrm{A}}, {F}_{\mathrm{B}} \gets \left\{{F}_{\mathrm{slip}}, {F}_{\mathrm{stick}}\right\}|_\mathrm{A\,slip}$\;
    }
    {   
        $\mathcal{S} =$ B slip\;
        ${F}_{\mathrm{A}}, {F}_{\mathrm{B}} \gets \left\{{F}_{\mathrm{slip}}, {F}_{\mathrm{stick}}\right\}|_\mathrm{B\,slip}$\;
    }
    \If{${F}_\mathrm{stick}^\mathrm{n}, {F}_\mathrm{stick}^\mathrm{t}, {M}_\mathrm{stick}^\mathrm{n}$ $\mathrm{satisfy}$ (\ref{equ::condition for trasition})}{
        $\mathcal{S} =$ two-finger slip\;\
        ${F}_{\mathrm{A}}, {F}_{\mathrm{B}} \gets$ Algorithm\_1 $(\boldsymbol{W}_{\mathrm{req}}, \mu_i, \boldsymbol{n}_i)$\;
    }

    \caption{Solving Force with the required wrench.}
    \label{alg::calculation of force}
\end{algorithm}
\DecMargin{1em}

% 基于前两个章节的分析，可以提出针对未知物体的多次抓取框架，如图所示。首先是第一次抓取，测量当前的接触力与接触力矩，并利用公式（1）确定当前物体所需的力旋量，最后通过算法3确定当前时刻物体所需抓取力，并基于该值设定机械手抓取力控制目标。在抬升过程中不断重复上述步骤，直到将物体抓取起来。抓取起来以后利用公式（2）测量物体重心未知与重力大小。在再抓取过程中，利用测量得到的物体参数直接预测物体抬升起来以后所需的力旋量，即公式（3）（4）。接着采用算法3确定抬升物体所需的最终抓取力，将抓取力控制目标直接设定为高于该值一定阈值，之后直接将物体抬升起来，从而使得再抓取过程不再需要不断谨慎反馈。

\begin{figure}[t]
    \centering
    \includegraphics[width = 1\linewidth]{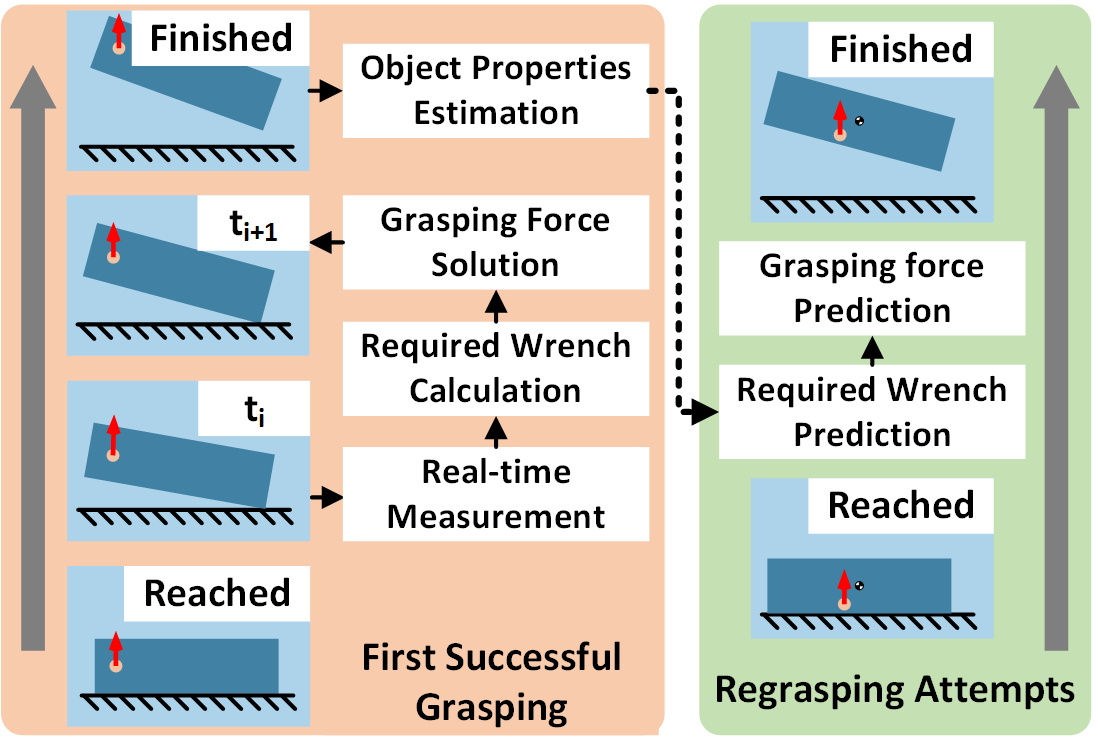}
    \caption{Grasping process for unknown objects.}
    \label{fig::grasping scenarios}
\end{figure}

Finally, a multiple grasping framework for an unknown object is proposed, as shown in Fig. \ref{fig::grasping scenarios}. Initially, during the first grasp, the current contact force and contact torque are measured, and equation (\ref{equ::required wrench})--(\ref{equ::require torque from feedback}) is used to determine the required force spinor for the object. The grasping force needed at that moment is then calculated using Algorithm (\ref{alg::calculation of force}), and the gripper's grasping force control target is set accordingly. These steps are repeated throughout the lifting process until the object is securely grasped. After the object is successfully lifted, equation (\ref{equ::measurement of G})--(\ref{equ::special solution of l}) is used to measure the object's unknown center of gravity and weight. During the regrasping process, the measured object parameters are used to directly predict the required wrench after the object is lifted, using equations (\ref{equ::require force from prediction}), (\ref{equ::require torque from prediction}). Algorithm (\ref{alg::calculation of force}) is then applied to determine the final grasping force needed to lift the object, and the grasping force control target is set to a threshold slightly higher than this value. The object is then lifted directly, eliminating the need for continuous cautious feedback during the regrasping process.

\section{Experiment}
% 引入新的实验。为三明治盒的抓取。桌面上摆放10个三明治盒，可能存在遮挡。预设抓取位置。
% 采用三种方法抓取。分别为保证效率的恒力抓取（效率差不多，但是抓取力过大导致失败，抓取力依据最开始的10个随机抓取位置而生成最理想的估计），需要补充说明不同位置的强度也不同，因此没法试着捏坏来确定上界（也就没法当破坏的力小于抓取力时，重新生成抓取位置）；保证成功率的反馈抓取；以及采用了我们的方法的抓取。最后给出三种方法的总成功率、总时间。

% 介绍实验平台，机械臂（抬升速度），运算平台（算力），给出图片
Four experiments were conducted to validate the effectiveness of the proposed method. First, we utilized two experiments to validate the accuracy and adaptability of the force prediction. Next, we integrated grasp force prediction into the grasp force control strategy and experimentally validated the effectiveness of this integration in improving grasping efficiency. Additionally, another experiment demonstrated the applicability of the prediction-integrated method across a variety of everyday objects. Finally, in a comparative demonstration, we evaluated our method against pure feedback-based approaches and another method that employs predefined constant grasping forces. This comparison confirmed the effectiveness of our approach in enhancing both grasping efficiency and success rates across multiple regrasping scenarios.

\begin{figure}[t]
    \centering
    \includegraphics[width = 0.6\linewidth]{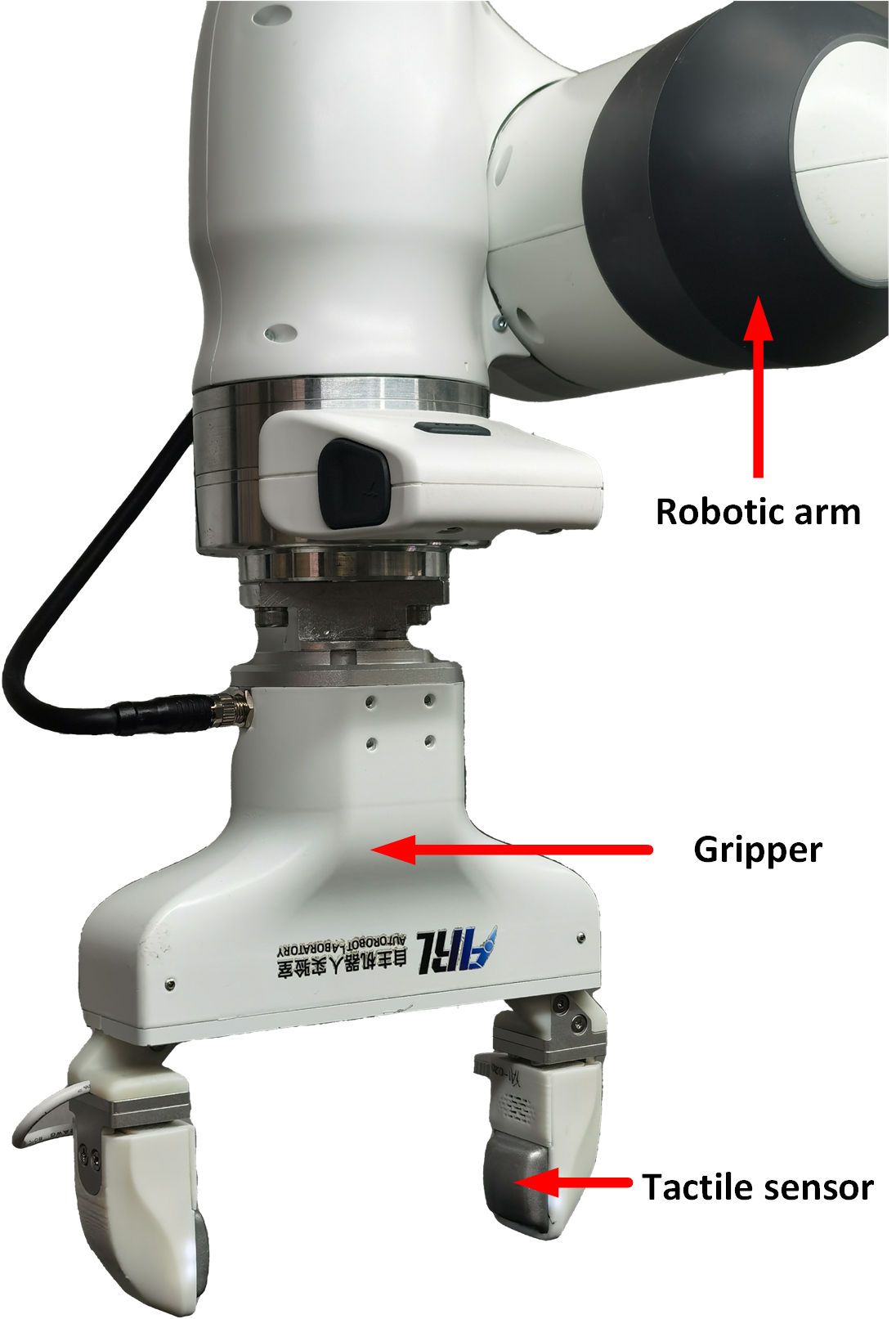}
    \caption{Experiment platform.}
    \label{fig::platform}
\end{figure}

The experimental setup is depicted in Fig. \ref{fig::platform}. The robotic arm, equipped with a gripper, was fitted with the Tac3D tactile sensor on its fingertips. The Tac3D sensor features a soft silicone surface and, based on reference \cite{nicholas1999Modeling}, is characterized by a contact coefficient $c=2.07$, a power-law equation exponent $\gamma=0.259$, and a pressure distribution coefficient $k=2$. Under the experimental conditions of this study, the maximum force measurement error for the sensor is 0.1 N, and the maximum torque measurement error is 1 N$\cdot$mm.

% 说明共同需要哪些信息：摩擦系数、表面倾角、接触面合力、合力矩；说明预测还需要重心，重心来源于反馈测量。
The feedback-based method and force prediction necessitate four types of information from the contact surface: the friction coefficient, surface normal, resultant force, and torque. For acquiring the resultant force and torque data from the contact surface, we opted for the Tac3D tactile sensor \cite{Zhang2023} (other 6D force-torque sensors can also be employed). 
% 给出表面倾角测量方法
Given the capability of Tac3D to measure the 3D deformation of the contact surface, it substitutes traditional vision-based methods to obtain the contact surface normal. Before lifting the object, the fingers are moved to slightly slide over the object while measuring the current tangential and normal forces at the contact surface. This data will be used to calculate the friction coefficient of the object.

\begin{figure}[t]
    \centering
    \includegraphics[width = 1\linewidth]{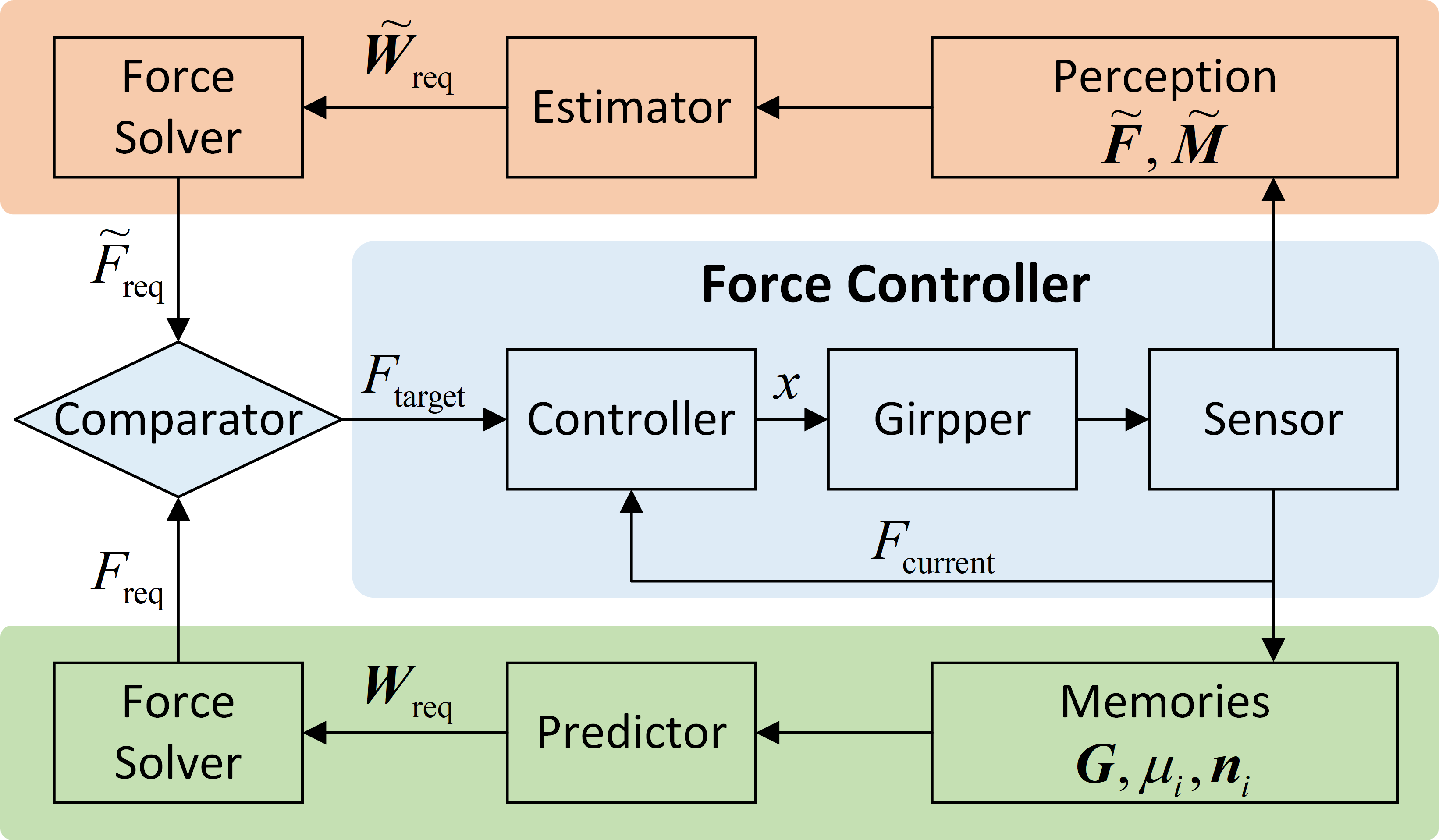}
    \caption{Gripping force control framework. The comparator controls whether to use force prediction. The controller takes the target force $F_\mathrm{target} $ as input and the distance of finger $x$ as the output, utilizing a well-tuned PID control.}
    \label{fig::control scheme}
\end{figure} 

The control diagram depicted in Fig. \ref{fig::control scheme} outlines the control scheme utilized in the experiments. The Force Controller, implementing conventional PID control, is utilized to control the gripper and attain the desired grasping force. The input for this controller is the target force $F_{\mathrm{target}}$, while the output is the distance $x$ between two fingers. The PID parameters maintain the same values across all experiments. The Comparator is responsible for determining whether to engage the predicted grasping force. Operating with a threshold of 10\%, its logic is as follows:
\begin{equation}
    \label{equ::comparator}
    F_{\mathrm{target}} = \begin{cases}
        1.1 \, \widetilde{F}_{\mathrm{req}}, &{\text{no prediction}}\\
        1.1 \, \max\left(\widetilde{F}_{\mathrm{req}}, F_{\mathrm{req}}\right), &{\text{use prediction.}}
    \end{cases}
\end{equation}

\subsection{Accuracy of Required Force Prediction}
\begin{figure}[!t]
    \centering
    \includegraphics[width = 1\linewidth]{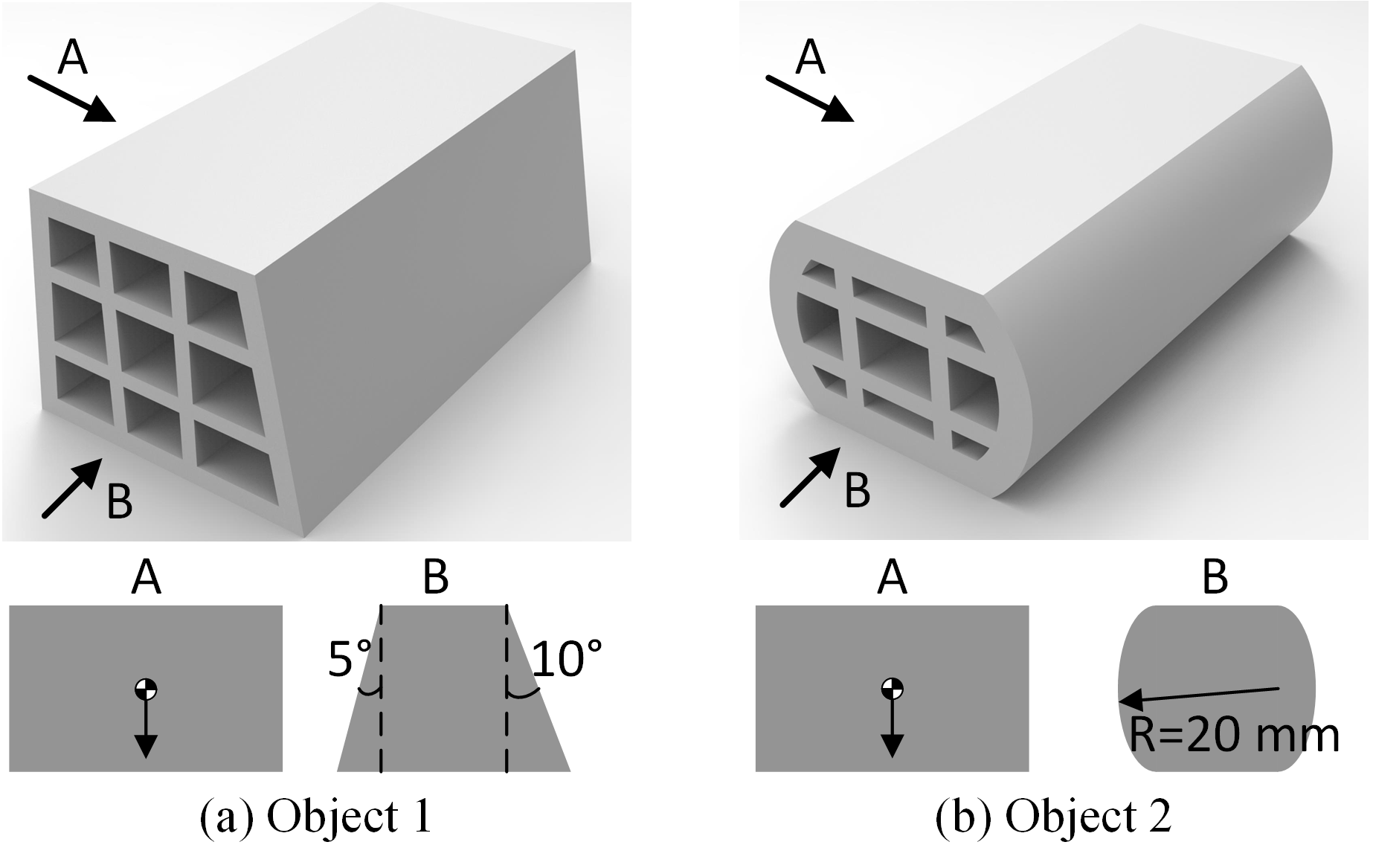}
    \caption{Two different grasping items: (a) Asymmetric quadrilateral prism and (b) Curved objects.}
    \label{fig::grasping objects}
\end{figure}

% 说明抓取物体的特点。抓取对象：非对称平面物体；对称曲面物体。
To comprehensively assess the accuracy of the minimum required grasping force prediction, we employed an asymmetrical trapezoidal column as the grasping object. Specifically, we selected positions that were deliberately offset from the center of mass to evaluate the applicability of the method in situations when torque is applied by the fingers. Additionally, we chose a curved object to grasp, as depicted in Fig. \ref{fig::grasping objects}. The curvature radius of Object 2 (20 mm) was set to be smaller than the curvature radius of the fingers (39 mm).

% 给出摩擦系数测量方法
To determine the friction coefficient, the following procedure was followed: (1) Apply a relatively small force to the object; (2) Move the finger nearly tangentially to induce slippage between the finger and the object; (3) Calculate the friction coefficient using the tangential and normal forces; (4) Repeat step (3) three times, averaging the results to obtain the measured friction coefficient. The relative slippage between fingers and objects can be very small, allowing the measurement of the coefficient of friction to be incorporated into each grip attempt. That is, in the subsequent grasping experiments, before grasping objects in subsequent experiments, a slight slippage between the fingers and the object is induced to measure the coefficient of friction.

% 给出实验方法，给出抓取力-姿态曲线并分析实验结果。分析实验结果。
Utilizing the method detailed in Algorithm (\ref{alg::calculation of force}), we calculated the minimum required grasping forces across various positions and postures, subsequently comparing them with actual values obtained through the following steps: (1) Measure the friction coefficient between the object and the fingers; (2) Initiate the grasping process by applying a predetermined initial grip force and lifting the object off the ground; (3) Gradually reduce the applied grasping force until the object undergoes a slow sliding motion; (4) Utilize feedback from the tactile sensor to determine the actual minimum required grasping force; (5) Repeat the aforementioned steps five times and vary the given grasping positions or postures for each iteration.

\begin{figure}[t]
    \centering
    \includegraphics[width = 1\linewidth]{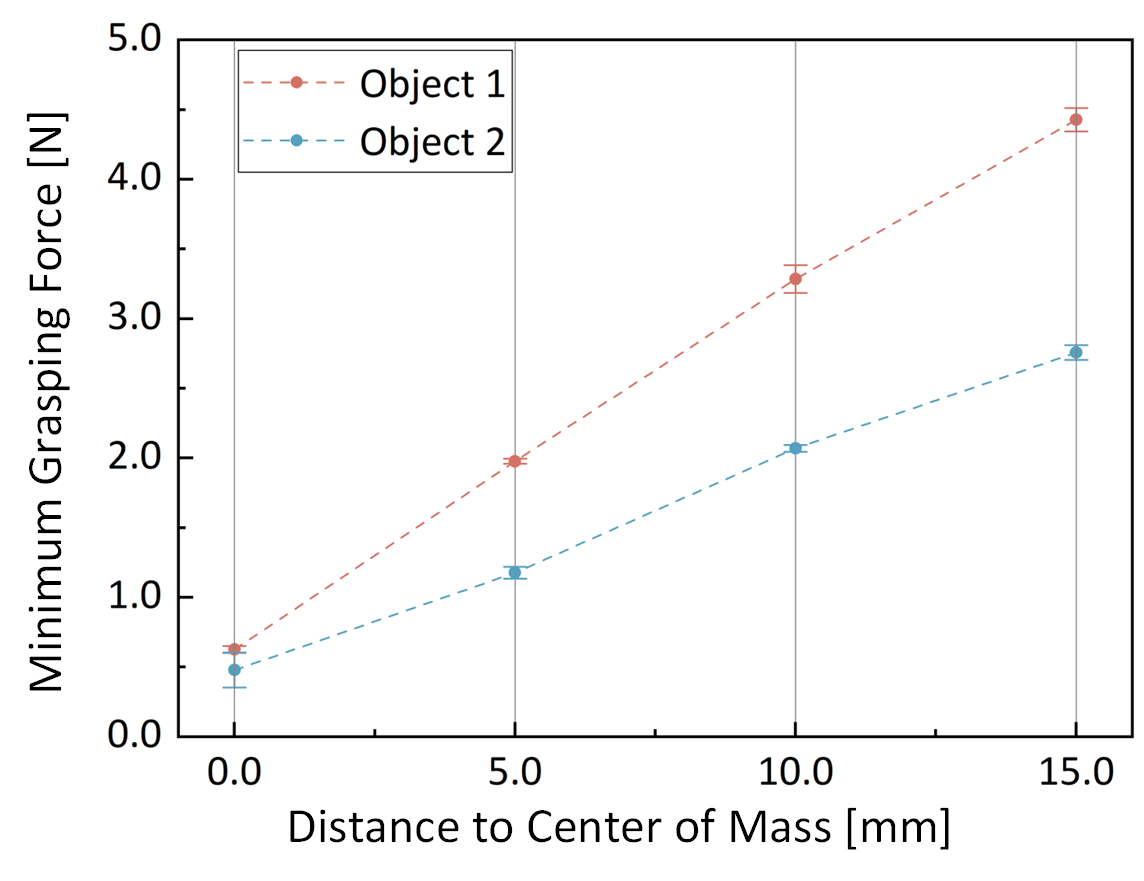}
    \caption{Results for different grasping positions. The actual experiment was conducted five times at each position, and the predicted grasping forces were compared. And the mean and standard deviation of the error were calculated.}
    \label{fig::force-result-predict-distance}
\end{figure}

\begin{figure}[t]
    \centering
    \includegraphics[width = 1\linewidth]{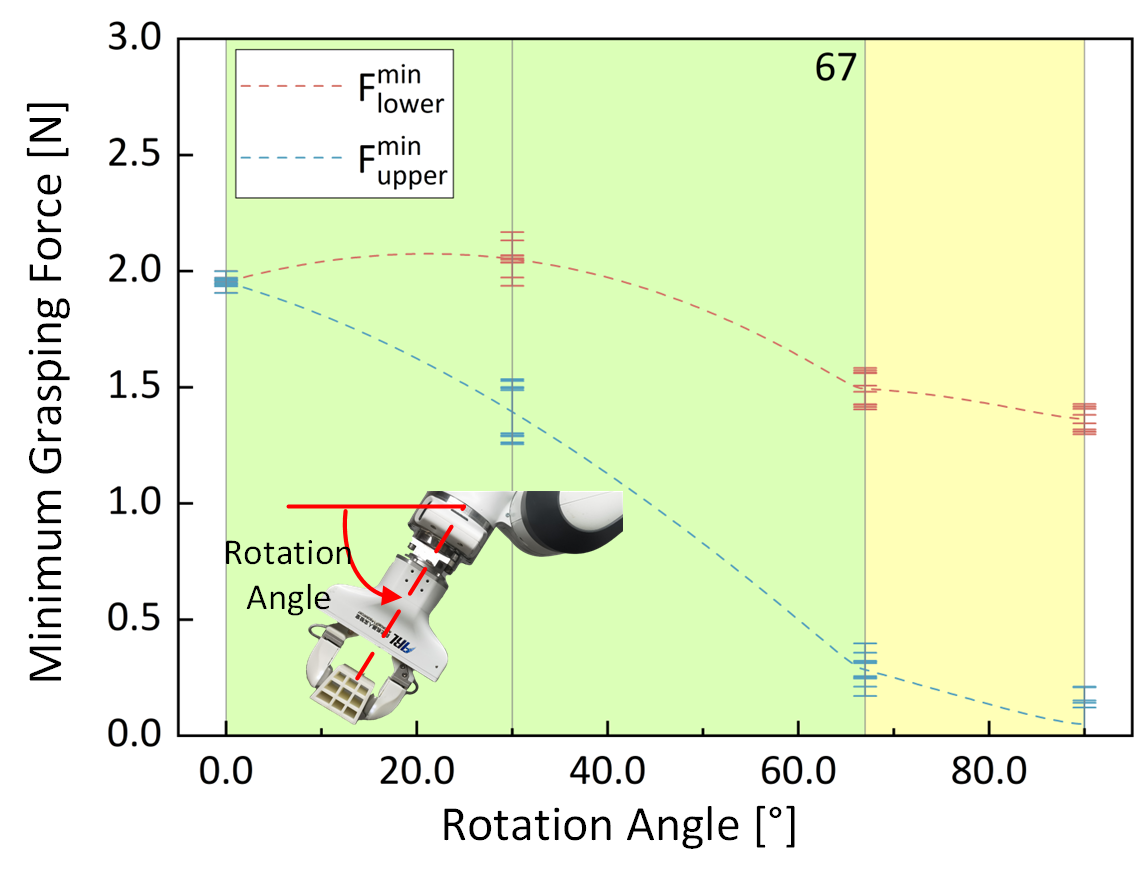}
    \caption{Results for different rotation angles. Dashed lines represent results obtained using predictions. The green background indicates the prediction of the two-finger slip state, while the yellow background indicates the one-finger slip state.}
    \label{fig::force-result-predict-rotate}
\end{figure}

% 分析两种物体的实验结果。
As shown in Fig. \ref{fig::force-result-predict-distance}, we first verified the accuracy of prediction on different position of Object 1. The error reaches its highest (0.13 N) when the distance to the center of mass is 10 mm. Additionally, to evaluate the error attributed to assuming a flat contact surface in the required force solution, experiments were conducted on a curved object (Object 2). The tendency of the results is similar to those of Object 1, showing a rapid increase in minimum required grasping force as the grasping position moved away from the center of mass. The error reaches its highest (0.18 N) when the distance to the center of mass is 0. The green background in the figure represents the two-finger slip state, which is determined by the proposed method in Section \uppercase\expandafter{\romannumeral5}-D.

Furthermore, validation of required force prediction accuracy under varied postures was performed on Object 1, as depicted in Fig. \ref{fig::force-result-predict-rotate}. With an increase in the rotation angle $\theta$, the grasping force for both fingers demonstrates a decreasing trend. The grasping force of the upper finger decreased at a faster rate. The yellow background indicates that at $\theta = 67^{\circ}$, the proposed method predicts the transition to a one-finger slip state. When the rotation angle is large, the upper finger tends to slip while the lower finger maintains stick to the object. The maximum prediction error for the minimum required grasping force was 0.16 N at $\theta = 90^{\circ}$ for the upper finger, and 0.12 N at $\theta = 30^{\circ}$ for the lower finger.

\subsection{Adaptability of Required Force Prediction}
\begin{figure}[t]
    \centering
    \includegraphics[width = 0.6\linewidth]{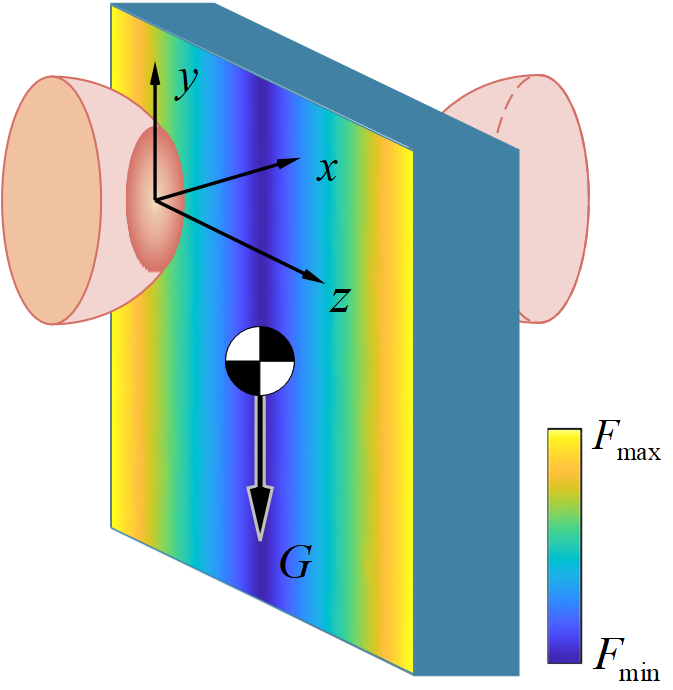}
    \caption{Schematic diagram of the distribution of the minimum grasping forces. To enhance visual clarity, only two-dimensional distributions are shown.}
    \label{fig::F map}
\end{figure}

% 给出不同物体的抓取力预测图，给出预测所需时间。
To evaluate the applicability and computational efficiency of force prediction, we conducted simulation experiments on various objects commonly found in daily life. The method was implemented on a machine equipped with an AMD R7-5800H CPU (3.2 GHz), utilizing MATLAB without parallel runtime or GPU acceleration. For our analysis, we leveraged 3DNet \cite{walter20123DNet}, a repository containing virtual object models extensively employed in simulations \cite{Newbury2022}. Specifically, our analysis encompassed four typical items: the banana representing normal objects, the donut representing objects with holes, the bowl representing concave objects, the hammer representing tool objects, and the Stanford bunny representing complex objects. In the experiment, a large number of grasping positions were uniformly sampled on different objects. Algorithm 3 was then used to calculate the minimum grasping force required for objects at these various positions. The total time required for all positions was recorded, and the average time required for each grasping position was calculated based on the number of positions, as shown in Table \ref{tab::simulation time}. For a specific grasping position, the required grasping force varies with changes in the object's posture. To maintain clarity, we kept the object's posture constant and set the grasping direction perpendicular to the $y$-$z$ plane, as illustrated in Fig. \ref{fig::F map}.

\begin{figure*}[t]
    \centering
    \includegraphics[width = 0.9\textwidth]{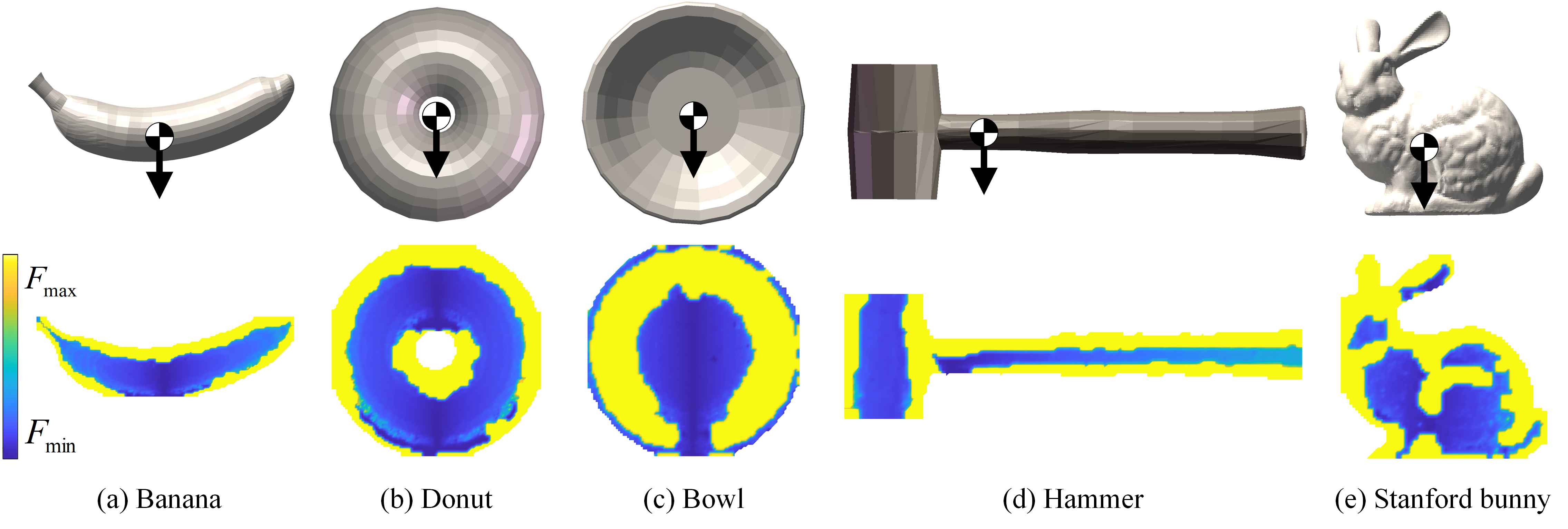}
    \caption{Results for (a) Banana, (b) Donut, (c) Bowl, (d) Hammer, and (e) Stanford bunny are displayed. The yellow region represents the grasping points where the grasping force exceeds ten times the minimum grasping force or the object cannot be grasped.}
    \label{fig::simulation results}
\end{figure*}

We predefined the centers of mass for the objects and set the friction coefficient to 0.5. The results are presented in Fig. \ref{fig::simulation results} and Table \ref{tab::simulation time}. There are two important trends observed. Firstly, the magnitude of the grasping force is closely related to the distance between the grasping position and the center of mass. As the grasping position moves farther away from the center of mass, there is a tendency for the grasping force to increase. The most force-saving grasping positions for different objects typically reside near their respective centers of mass.

Secondly, the magnitude of the grasping force is also influenced by the orientation of the contact surface normal. Taking the bowl as an example, For the grasping points near the edge of the bowl, even when they are far from the center of mass, the grasping force remains relatively low. This is because, compared to the inclined surface normal of the wall of the bowl, the flat platform at the mouth of the bowl is more conducive to the gripper exerting grasping force. At the same time, it is noticeable that the grasping force below the center of mass is relatively low compared to positions above the center of mass, indicating that the surface normal of the object being gripped does not necessarily need to be perfectly parallel to the gripping direction, but rather should have a certain angle with the gripping direction. This allows for better resistance against gravity for both normal and tangential forces at the contact point, potentially leading to a more force-efficient grasping position or posture.

\begin{table}[t]%调节图片位置，h：浮动；t：顶部；b:底部；p：当前位置
    \centering
    \caption{Solution Time of Different Objects}
    \label{tab::simulation time}  
    \begin{tabular}{cccc}%表格中的数据居中，c的个数为表格的列数	
        \toprule
        \makebox[0.1\linewidth][c]{Objects} &
        \makebox[0.1\linewidth][c]{Points} & 
        \makebox[0.25\linewidth][c]{Total Time [s]} &  
        \makebox[0.25\linewidth][c]{Average Time [ms]}\\
        \midrule
        Banana   & 5122 & 134.3  & 26.2\\
        Donut    & 8700 & 223.6  & 25.7\\
        Bowl     & 7474 & 197.1  & 26.4\\
        Hammer     & 4141 & 116.2  & 28.1\\
        Bunny     & 4774 & 136.0  & 28.5\\
        \bottomrule
    \end{tabular}
\end{table}

\subsection{Validation of the efficiency-improving method}
% 验证抓取的效率提高
To assess the effectiveness of required force prediction in enhancing grasping efficiency, we conducted grasping experiments using Object 1 at various grasping positions. In these experiments, an iron block was added to Object 1 to obscure its center of mass in the beginning. The experimental procedure followed the illustration in Fig. \ref{fig::grasping scenarios}. 

For the first grasping of the object, we employed a feedback-based force control strategy outlined as follows: (1) Measured the friction coefficient between the object and the fingers; (2) Initiated the grasp using a predetermined initial force; (3) Lifted the object at a specified lifting speed while concurrently measuring the current external wrench, using equations (\ref{equ::require force from feedback}) and (\ref{equ::require torque from feedback}); (4) Utilized the method described in Algorithm (\ref{alg::calculation of force}) to determine the current minimum required grasping force; (5) Controlled the grasping force to be higher than the minimum required grasping force by a certain threshold (set at 10\%); (6) Continued lifting the object and adjusting the grasping force until it was successfully lifted off the ground; (7) Measured the center of mass of the object using the method described in Section \uppercase\expandafter{\romannumeral4}-C. 

Various lifting speeds were employed in the experiment, and the maximum lifting speed $v_\mathrm{max}$ capable of successfully grasping the object was established. Throughout the experiment, data on the minimum required grasping force obtained from sensor feedback, the target grasping force set for the gripper, and the actual grasping force exerted by the gripper were recorded. 

\begin{figure*}[t]
    \centering
    \includegraphics[width = 1\textwidth]{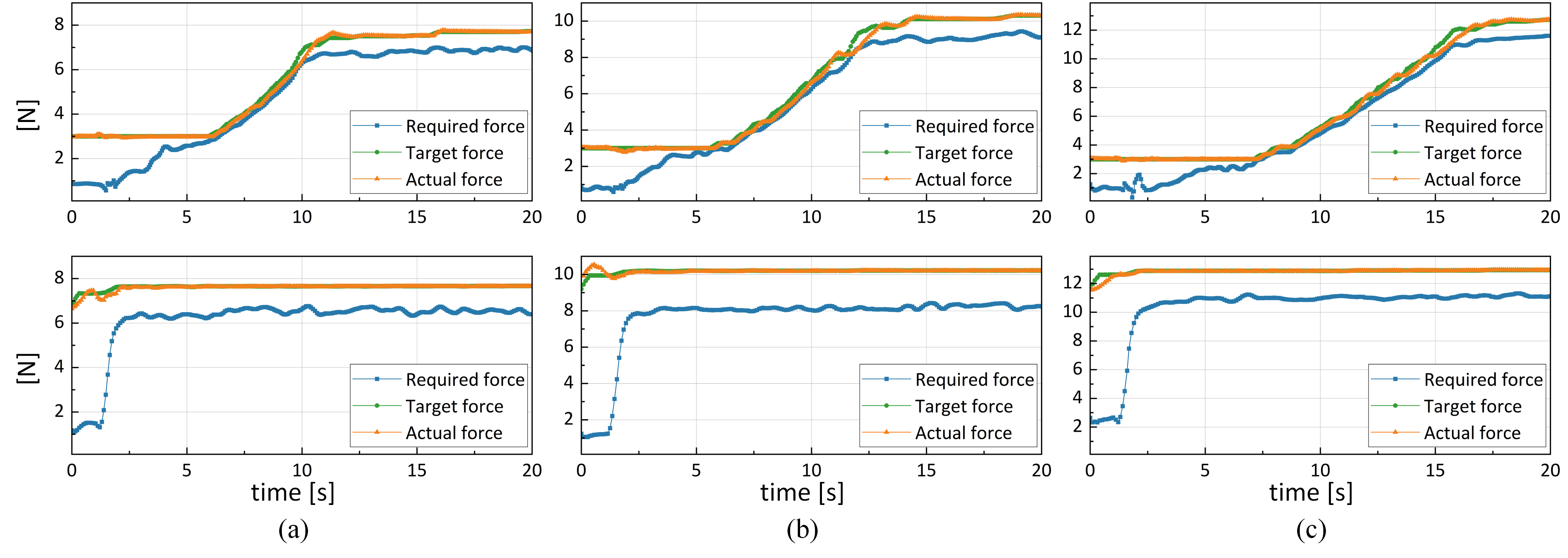}
    \caption{Gripping force control curves at different positions, each at a distance of (a) 10 mm, (b) 15 mm, and (c) 20 mm from the center of mass. The top row depicts results without employing prediction, while the bottom row illustrates results when utilizing prediction.}
    \label{fig::result of different l}
\end{figure*}

After determining the center of mass, the minimum required grasping force for a successful grasp was directly predicted. Using this value, we adjusted the initial grasping force. Subsequently, the object was lifted at a speed $v'_\mathrm{max} = 10 \, v_\mathrm{max}$. Similarly, the required grasping force, target grasping force, and actual grasping force were recorded. Our experiments were conducted at three different distances from the geometric center of the object. The results are illustrated in Fig. \ref{fig::result of different l}.

The experiment measured a lifting speed of $v_\mathrm{max}=1 \, \mathrm{mm/s}$, thereby setting $v'_\mathrm{max} = 10 \, v_\mathrm{max} = 10 \, \mathrm{mm/s}$. As the grasping time prolongs with the object continuously being lifted, the required force exhibits a steady increase. Correspondingly, both the target force and actual force rise proportionately until the object is successfully lifted. The results of the feedback-based force control method without prediction are shown in the top row of Fig. \ref{fig::result of different l}. When the speed reaches the maximum successful grasping speed $v_\mathrm{max}$, the required force becomes close to the actual force in the lifting phase. This alignment suggests the object is prone to sliding, ultimately leading to grasping failure. 

On the other hand, when adjusting the initial grasping force based on the experimentally obtained center of mass measurement, even with an escalated lifting speed at $v'_\mathrm{max}=10 \, v_\mathrm{max}$, the actual grasping force consistently remains above a threshold higher than the minimum required grasping force. This assurance ensures the object remains undamaged and prevents slippage, facilitating stable and efficient grasping. It is observed that the threshold exceeds the preset value of 10 \%. This deviation could be due to the challenge of maintaining the object posture unchanged within the gripper when the predictive method is not employed. Although the method ensures no slippage between the fingers and the object, due to the tendency to slip and the elasticity of the soft fingers, the object is prone to rotate slightly under the influence of gravity. This rotation leads to a reduction in the torque exerted by gravity on the object, resulting in the gravity measurement error. This discrepancy is also evident in the force curves observed when employing the predictive method (lower row in Fig. \ref{fig::result of different l}): the required grasping forces exhibit varying degrees of increase in the beginning.

\subsection{Adaptability of the efficiency-improving method}
\begin{figure*}[t]
    \centering
    \includegraphics[width = 1\textwidth]{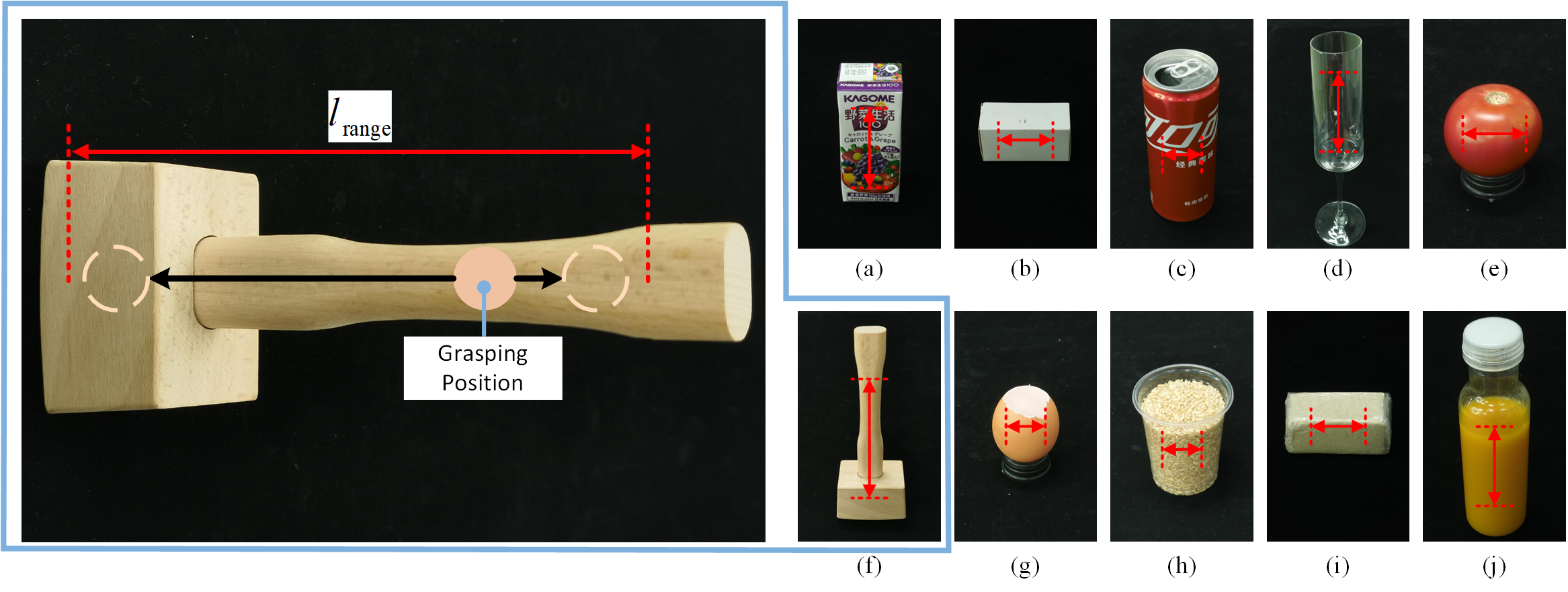}
    \caption{Different objects from daily life and their relative postures to the gripper in the experiment. Each object will be randomly grasped within the grasping range $l_\mathrm{range}$, with ten grasping attempts made for each object.}
    \label{fig::grasp items}
\end{figure*}

% 分别在以下物体上验证方法成功率（这里的成功定义为不滑移+不破坏物体）:豆腐、剥皮香蕉、成熟的西红柿、橡皮泥、半个鸡蛋壳、空的铝罐、带水的瓶子（横着抓，说明方法适用性有限）、红酒杯、锤子、网球、偏心纸盒、板擦
% 一张大图放不同物体的照片
To assess the adaptability of the method across various objects, we conducted grasping experiments on 10 different natural objects depicted in Fig. \ref{fig::grasp items}. The maximum grasping force exerted by the gripper was set to 15 N.

% 说明实验流程, 需要试试不同物体的初始抓取力。
For each object, the experimental procedure was as follows: (1) Measure the friction coefficient between the object and the fingers. (2) Employ the grasping strategy depicted in Fig. (\ref{fig::grasping scenarios}) by continuously sensing the required force and moment during the “lift” phase. Adjust the grasping force until the object is successfully lifted. The initial force is set to 2 N, and the lifting speed is set to 1 mm/s. The safety threshold is established at 10\% higher than the desired grasping force. (3) Measure the gravity and the center of mass position using equations (\ref{equ::measurement of G})--(\ref{equ::special solution of l}) and place the object back in its original position. (4) Generate the next grasping position randomly and predict the required grasping force using the method outlined in Algorithm (\ref{alg::calculation of force}). (5) Adjust the initial force to the predicted value and lift at a speed of 10 mm/s. (6) Repeat steps (4) and (5) a total of 10 times, and calculate the success rate. Due to the small differences in friction coefficients for the same object at different positions, we measured the friction coefficient only in the first grasping attempt and applied this measured value in subsequent grasps at different positions.

% 随机生成抓取位置需要说明结合物体大小，随机生成的位置也只是平移
In the experiment, the randomly generated grasping positions are depicted in Fig. \ref{fig::grasp items}, and the robotic arm moves in the z-direction of the ground coordinate system. The range of random positions $l_{\mathrm{range}}$ is initially set based on the size of the object. Moreover, considering potential object localization errors resulting from vision-based object localization, the initial object position remains consistent across multiple grasping attempts. The relative position between the gripper and the object is acquired through feedback from the robotic arm.

% 实验结果分析
The results are presented in Table \ref{tab::grasp results}, where $F_{\mathrm{max}}$ denotes the maximum grasping force without crushing the object, and $l_{\mathrm{range}}$ denotes the range of randomly selected grasping position, with the center of mass as the origin. In the experiments, the required grasping force notably increases as the distance from the center of mass increases.

For both the champagne glass, the hammer, and the half-bottled juice, the success rate of grasping is influenced by the grasping range $l_{\mathrm{range}}$. When the grasping position is too far away from the center of mass, the required grasping force may surpass the maximum capacity of the gripper, resulting in slip failures. Restricting the grasping range near the center of mass ensures that the required force for successful grasping remains within the maximum capability of the gripper, leading to a notable improvement in the success rate.

Regarding the tofu, achieving successful grasping demands a higher precision in the grasping range due to its low tolerance for maximum grasping force. Without constraining the grasping range, there is a high likelihood of crushing the object. The revised $l_{\mathrm{range}}$ is determined using the algorithms proposed in this study, which computes the distance between the grasping position and the center of mass when the grasping force reaches the limit of the gripper or of the object, thereby establishing the range boundaries for random grasping positions.

For the half-bottled juice, even with modifications made to the grasping range $l_{\mathrm{range}}$, there remains a probability of failure (20\%). This probability arises because the internal juice can cause a shift in the center of mass with variations in the grasping position, resulting in a margin of error in predicting the required grasping force.

% 一个表格放不同物体的成功率，其中存在抓取力上限的需要标注，最好有失败案例。附上偏心距离范围、摩擦系数。
\begin{table}[t]%调节图片位置，h：浮动；t：顶部；b:底部；p：当前位置
    \centering
    \caption{Grasping results of different objects.}
    \label{tab::grasp results}
    %\begin{threeparttable}
    \begin{tabular}{lcccc}%表格中的数据居中，c的个数为表格的列数	
        \toprule
        \makebox[0.1\linewidth][c]{Objects} &
        \makebox[0.04\linewidth][c]{$\mu$} & 
        \makebox[0.12\linewidth][c]{\makecell[c]{$F_{\mathrm{max}}$ [N]}} &  
        \makebox[0.12\linewidth][c]{\makecell[c]{$l_{\mathrm{range}}$ [mm]}} &
        \makebox[0.17\linewidth][c]{Success Rate}\\
        \midrule
        Carton pack juice   & 1.17 & 12.4   & [-40,40]  & 100\% \\
        Box with contents   & 1.04 & $>15$  & [-20,40]  & 100\% \\
        Empty can           & 0.29 & 6.2    & [-15,15]  & 100\% \\
        Ripe tomato         & 1.16 & 11.2   & [-15,15]  & 100\% \\
        Half egg shelf      & 0.37 & 5.6    & [-7.5,7.5]  & 100\% \\
        Plastic cup of rice & 0.48 & 9.5    & [-7.5,7.5]  & 100\% \\
        \cmidrule{0-2}
        \multirow{2}{*}{Champagne glass} & \multirow{2}{*}{0.70} & \multirow{2}{*}{$>15$} & [80,160]  & 30\% \\
        \multirow{2}{*}{}          & \multirow{2}{*}{} & \multirow{2}{*}{} & [80,90]  & 100\% \\
        \cmidrule{0-2}
        \multirow{2}{*}{Hammer}          & \multirow{2}{*}{1.14} & \multirow{2}{*}{$>15$} & [-15,160]  & 20\% \\
        \multirow{2}{*}{}          & \multirow{2}{*}{} & \multirow{2}{*}{} & [-15,15]  & 100\% \\
        \cmidrule{0-2}
        \multirow{2}{*}{Tofu}            & \multirow{2}{*}{0.70} & \multirow{2}{*}{3.2}    & [-20,20]  & 10\% \\
        \multirow{2}{*}{}          & \multirow{2}{*}{} & \multirow{2}{*}{} & [-3,3]  & 100\% \\
        \cmidrule{0-2}
        \multirow{2}{*}{Half-bottled juice}   & \multirow{2}{*}{2.03} & \multirow{2}{*}{$>15$}  & [-20,80]  & 40\% \\
        \multirow{2}{*}{}   & \multirow{2}{*}{} & \multirow{2}{*}{}  & [-20,50]  & 80\% \\
        \bottomrule
    \end{tabular}

    %\begin{tablenotes}
        % \footnotesize
    %    \footnotetext{$F_{\mathrm{max}}$: the maximum grasping force without crushing the object.}
    %    \footnotetext{$l_{\mathrm{c}}$: distance to the object's center of mass.}
    %\end{tablenotes}
    %\end{threeparttable}
\end{table}

\subsection{Demonstration Experiment}
% 展示性实验的实验流程如图所示。机械手依次完成三个场景的物体抓取，并分别将香蕉和蛋糕盒放在指定位置内，实验中统计不同场景所需的时间。分别采用以下三种方式控制抓取力：（1）预设恒定抓取力，实验中恒力设定为物体所能承受的最大抓取力的一半；（2）基于纯反馈的方法；（3）将预测融入到反馈控制过程中。

% 最终实验结果如表所示。预设且恒定的抓取力后，在每次抓取的抓取效率较高，但由于抓取力不一定满足需求，因此成功率无法保证；基于纯反馈的方法成功率相对较高，但是效率受限。最后，当融入预测后，在后续再抓取过程中效率相较于基于反馈的方法有了显著的提高。同时，基于预测的方法可以在抓取前预测所选抓取位置是否可以将物体抓取起来，即所需抓取力小于机械手最大输出力且小于物体最大破坏力，从而避免了失败的抓取尝试。需要补充，这里的预测是基于物体处处摩擦系数一致的假设，当摩擦系数不一致时，则需要通过初步接触物体等方式来测量摩擦系数。

\begin{figure}[t]
    \centering
    \includegraphics[width = 1\linewidth]{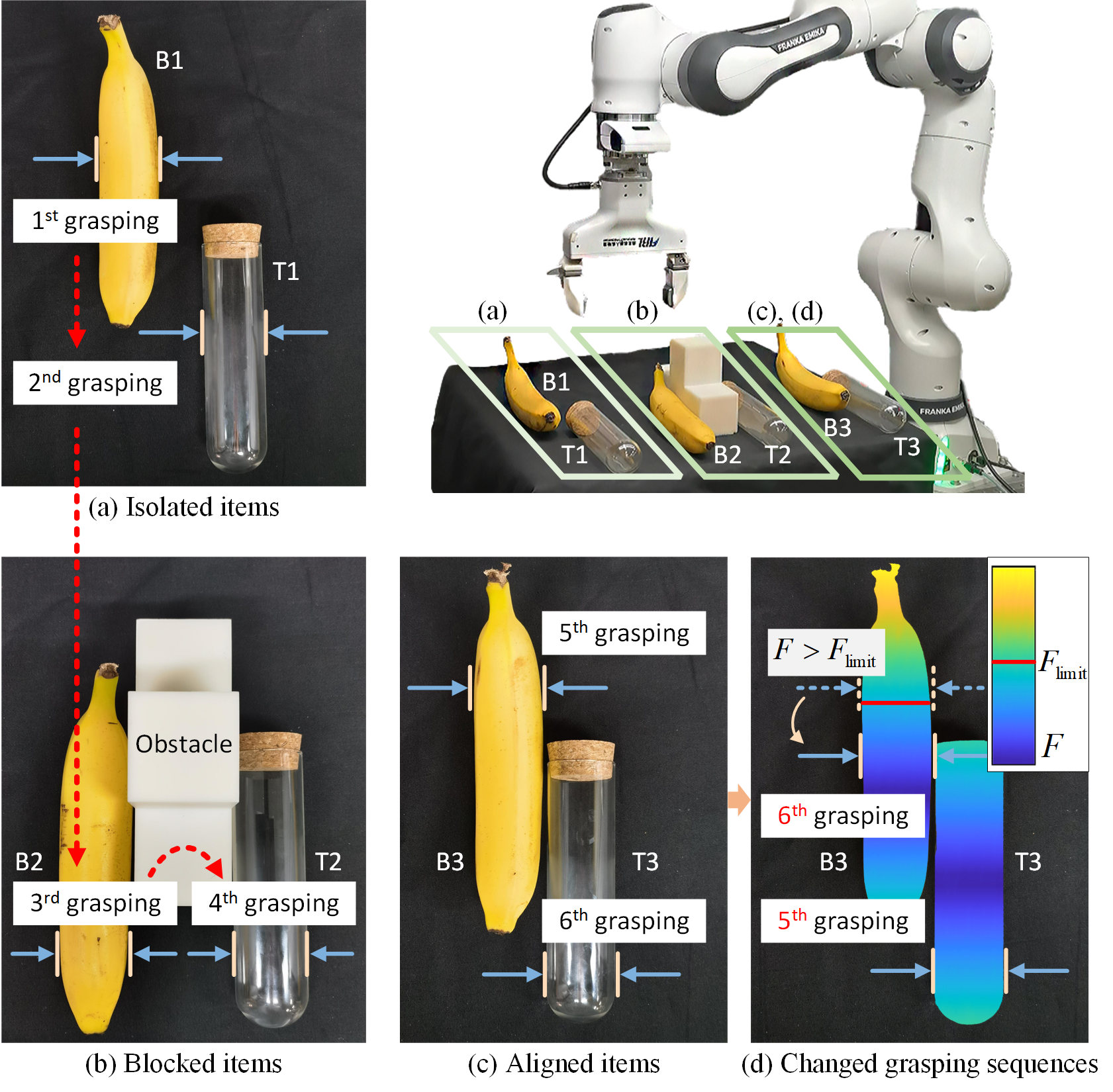}
    \caption{The grasping positions and sequence for different scenarios.}
    \label{fig::demonstration}
\end{figure}

The experimental procedure for the demonstration experiments is illustrated in Fig. \ref{fig::demonstration}. The maximum safe grasping force $F_\mathrm{limit}$ is set at 10 N. The gripper sequentially grasp the objects in three different scenarios, and place them in designated locations, while recording the time required for each scenario. The grasping forces are controlled using three methods, respectively: (1) a predefined constant grasping force (5 N), set at half of the maximum safe force the object can withstand; (2) a pure feedback-based method; and (3) an approach that incorporates prediction into the feedback control process.

The final experimental results are summarized in Tab. \ref{tab::demonstration results}. With a predefined and constant grasping force, grasping efficiency is relatively high; however, since the force may not always meet the requirements, the success rate is not guaranteed. The pure feedback method yields a higher success rate but is limited in efficiency. In contrast, integrating prediction significantly enhances efficiency during subsequent regrasping compared to the feedback-based method. Additionally, the predictive approach allows for the assessment of whether the chosen grasping position can successfully grasp the object—ensuring that the required grasping force is less than both the maximum output force of the robotic hand and the object's maximum tolerable force, thereby preventing failed grasping attempts. It is important to note that this prediction is based on the assumption of consistent friction coefficients across the object; if friction coefficients vary, it is necessary to measure the friction coefficients in advance.

\begin{table}[t]%调节图片位置，h：浮动；t：顶部；b:底部；p：当前位置
    \centering
    \caption{Grasping Time of Different Methods}
    \label{tab::demonstration results}  
    \begin{tabular}{c|cc|cc|cc}%表格中的数据居中，c的个数为表格的列数	
        \toprule
        \multirow{2}{*}{Methods} &
        \multicolumn{2}{c|}{Isolated [s]}& 
        \multicolumn{2}{c|}{Blocked [s]}& 
        \multicolumn{2}{c}{Aligned [s]}\\
        \multirow{2}{*}{} & B1 & T1 & B2 & T2 & B3 & T3 \\
        \midrule
        PF & 5.8 & 5.8 & Failed & 5.6 & Failed & 6.4\\
        FB & 26.4 & 26.2 & 27.0 & 28.3 & Falied & 31.1\\
        Ours & 26.2 & 26.3 & 5.5 & 5.7 & 6.2 & 6.3\\
        \bottomrule
    \end{tabular}
    \begin{tablenotes}
		\item PF: Pre-set force;  FB: Feedback-based force control
    \end{tablenotes}
\end{table}

\section{Discussion}
% 说明实验结果、意义，误差来源，局限
By adjusting the grasping force using predictions, the experimental findings suggest that stable grasping on various everyday objects remains achievable even when increasing the lifting speed to 10 times its original limit. After integrating force prediction, the grasping time has been reduced from 27.0 s to 6.3 s. Though assuming that the contact surface needs to be approximated as flat, the method still demonstrates good precision (0.18 N) when applied to surfaces with large curvature (curvature radius 20 mm $<$ finger curvature radius 39 mm). Moreover, the predictive algorithm for determining the required grasping force demonstrates robust applicability, effectively adapting to diverse and complex objects with varying shapes (with an average computation time of less than 29 ms). The results of the demonstration experiments show that incorporating prediction into the feedback process can reduce grasping time from 26.2 s to 6.3 s.

% 分析精度误差
One significant source of accuracy error in determining the final force arises from assuming a central symmetric distribution of the pressure within the contact surface, as outlined in equation (\ref{equ::pressure distribution}). The magnitude of this distributed normal force impacts both the resultant force and moment applied by the fingers to the object, thereby influencing the accuracy of grasping force prediction. Force distribution can be influenced by factors such as contact surface morphology and finger shape. To attain more precise force distribution characteristics, it would be beneficial to use sensors capable of measuring force distribution, such as the Tac3D \cite{Zhang2023} used in this study, or other deformable sensors like GelSight \cite{Abad2020}.

% 分析适用性，针对存在棱角、重心变化的物体
Gripping force prediction relies on measuring the center of mass. For objects where the center of mass significantly shifts, such as containers with flowing liquids, the effectiveness of this method may be limited. Raising the safety threshold could be one way to make sure that the grasping force is still sufficient to meet the necessary requirements, even in cases when there is a certain amount of center of mass displacement. An alternative and efficient strategy could entail a heightened dependence on real-time measurements to consistently update the position of the center of mass.

% 说明方法局限，引入抓取位形选取工作
The aim of predicting the minimum required grasping force is to prevent object slippage while avoiding damage. However, for heavy objects, the required grasping force might exceed the maximum capacity of the robotic hand. In such instances, it becomes essential to assess and filter grasping positions based on whether the required grasping force exceeds the limit of the gripper. Additionally, for certain off-center positions of delicate objects like tofu, the required grasping force might exceed the object's tolerance, and the required grasping force might surpass the object's tolerance level, making successful grasping unfeasible without risking damage. To mitigate the risk of object damage, filtering grasping positions based on whether the grasping force exceeds the object's tolerance limit can be a promising approach.

\section{Conclusion}
% 总结全文工作，后续研究规划，说明潜在用途，对学术界的价值
Our method significantly improves grasping efficiency in handling unknown objects after the first successful grasping. By accurately predicting the required grasping force for grasping the object, our approach allows for quicker lifting of objects. In the dynamic process of grasping, interactions with the environment pose challenges in determining the appropriate grasping force, requiring continuous perception of the current state and adjustment of the grasping force. Our research reveals predicting the final required grasping force significantly reduces real-time perception of external forces, thereby further optimizing grasping efficiency.

Our approach outlines characteristics and methods for calculating and predicting the wrench required on the contact surface. This prediction assists in determining the current required grasping force. Once the finger material is given, key grasping information necessary for predicting the required grasping force includes the center of mass position, friction coefficient, and orientation of the contact surface. Moreover, we present a methodology for determining the minimum required grasping force for a parallel gripper considering soft fingertips. Establishing this minimum grasping force threshold holds significance in two-finger grasping as it serves to prevent object slippage. Additionally, this threshold stands as a novel evaluation metric used to assess grasping performance.

The success criterion for grasping in this study primarily is preventing the object from dropping, which further requires avoiding macro sliding between the object and the fingers. However, there are certain specific macro sliding conditions, such as rotational slippage, where the object may not necessarily fall from the hand but instead pivots within the grasp. Future research endeavors will further investigate the predictability of this rotational movement and its potential impact on the overall grasping process.

{\appendix[Limit Surface Theory]
When deformation is not negligible, a soft contact model should be considered, where the contact surface is no longer a single point. According to the theory and the experimental validation, the soft contact model can be categorized into the time-dependent model and the time-independent model. For the time-independent model, Xydas and Kao developed the power-law equation \cite{nicholas1999Modeling}:
\begin{equation}
    \label{equ::power law theory}
    R=c F_\mathrm{n}^{\gamma},
\end{equation}
where $R$ is the radius of the contact surface and $F_\mathrm{n}$ is the normal force and $c$ is the coefficient depending on the size, shape of the fingertip, and material properties. $\gamma=n/(2n+1)$ is the exponent of the power-law equation with $n\in \left[0,1\right] $ being the stress exponent for elastic materials. Apart from the time-independent model, another group is the time-dependent model or viscoelastic model. These models consider the creep and relaxation phenomena in soft materials \cite{Tiezzi2007}. Compared with the time-independent models, the viscoelastic models are more time-consuming.

The distribution of the pressure satisfies the following equation:
\begin{equation}
    \label{equ::pressure distribution}
    P(r) = C_{k} \frac{F_\mathrm{n}}{\pi R^{2}}\left[1-\left(\frac{r}{R}\right)^{k}\right]^{\frac{1}{k}},
\end{equation}
where $k$ relates to the material of the fingertip and determines the shape of the pressure profile. $C_{k}$ is used to adjust the pressure distribution to satisfy the equilibrium condition, where the integration of pressure over the contact surface equals the normal force $N$. $C_{k}$ can be obtained as
\begin{equation}
    \label{equ::Ck}
    C_{k}=\frac{3}{2} \frac{k \Gamma\left(\frac{3}{k}\right)}{\Gamma\left(\frac{1}{k}\right) \Gamma\left(\frac{2}{k}\right)}
\end{equation}
where $\Gamma(\cdot)$ is the Gamma function. The coefficient with $k\geq2$ are more common used pressure distributions.

\begin{figure}[t]
    \centering
    \includegraphics[width = 0.6\linewidth]{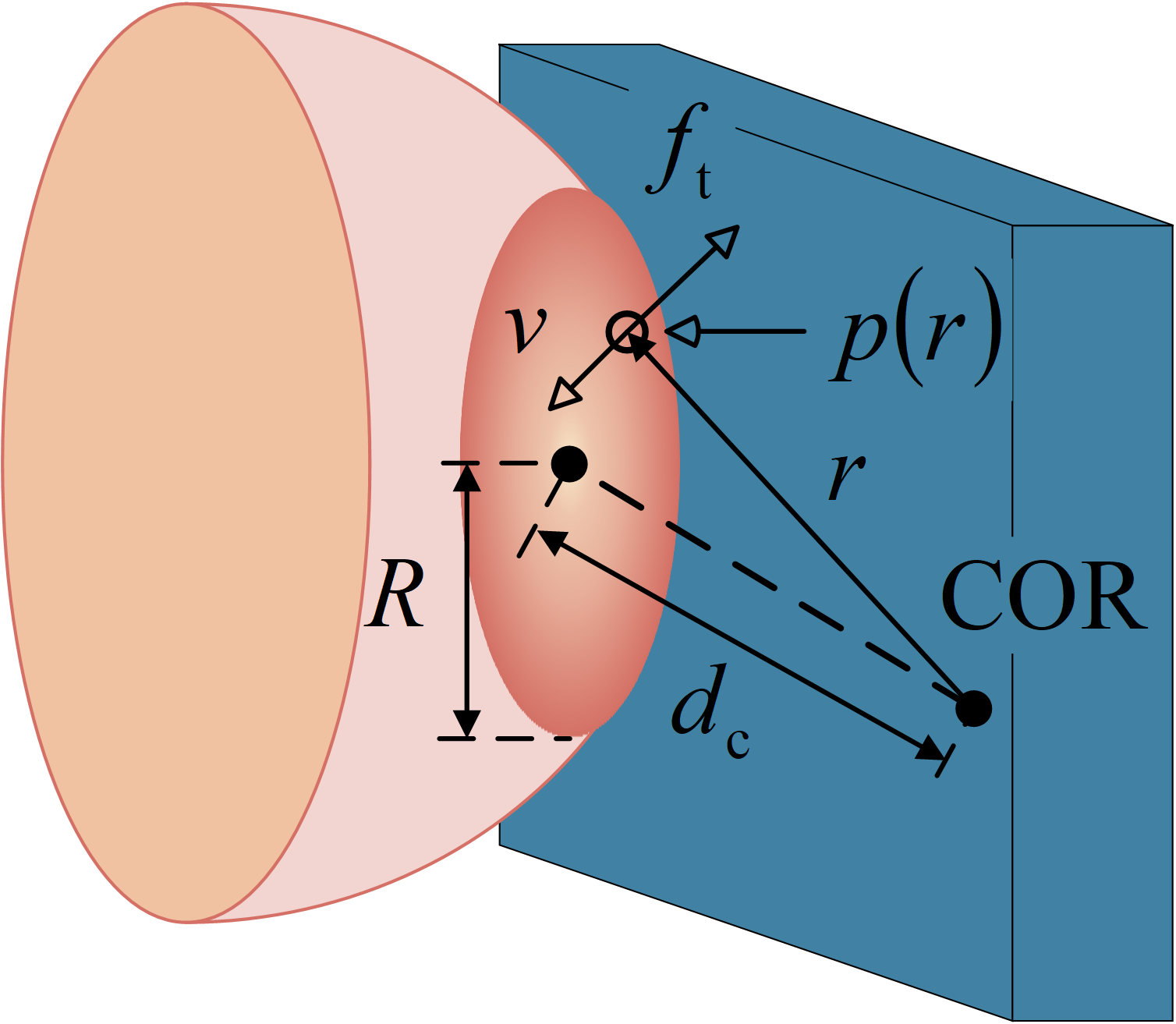}
    \caption{The slip state of the soft finger contact, along with its center of rotation, determines the direction of the distributed tangential forces denoted as $f_\mathrm{t}$.}
    \label{fig::center of rotation}
\end{figure} 

When the fingertip slips relative to the object, there exists a center of rotation (COR) \cite{s.1989Limit}. All points on the contact surface of the finger rotate around the COR, with the axis of rotation perpendicular to the normal direction of the contact surface, as shown in Fig. \ref{fig::center of rotation}. The direction of frictional force is inverted to the direction of relative motion and the tangential forces over the entire contact surface can be obtained by integrating the frictional forces over the infinitesimal surfaces:
\begin{equation}
    \label{equ::resultant tangential force}
    F_{\mathrm{t}}=\int \mu \frac{d \cos \theta-d_\mathrm{c}}{\sqrt{d^{2}+d_\mathrm{c}^{2}-2 d d_\mathrm{c} \cos \theta}} P(r) d A,
\end{equation}
where $\mu$ is the friction coefficient of the contact surface. $d$ is the distance between the COR and the infinitesimal surface. Similarly, the normal moment can be obtained:
\begin{equation}
    \label{equ::resultant normal moment}
    M_{\mathrm{n}}=\int \mu \frac{d^{2}-d d_\mathrm{c} \cos \theta}{\sqrt{d^{2}+d_\mathrm{c}^{2}-2 d d_\mathrm{c} \cos \theta}} P(r) d A.
\end{equation} 

The joint (\ref{equ::power law theory})--(\ref{equ::resultant normal moment}) leads to the following conclusion: the tangential force and the normal moment are approximately distributed on an ellipse.
\begin{equation}
    \label{equ::limit surface}
    \frac{F_\mathrm{t}^{2}}{F_\mathrm{max}^{2}}+\frac{M_\mathrm{n}^{2}}{M_\mathrm{max}^{2}}=1,
\end{equation}
where $F_\mathrm{max}=F_\mathrm{t} |_{d_\mathrm{c}=\to \infty}=\mu F_\mathrm{n}$ and $M_\mathrm{max}=M_\mathrm{n} |_{d_\mathrm{c}=0} $ are the maximum tangential force and normal moment under the force $F_\mathrm{n}$.
}


%\begin{thebibliography}{1}
\bibliographystyle{IEEEtran}
\bibliography{ref}

\vfill

\end{document}